%% file: main.tex
\documentclass{article}
\usepackage{iclr2027_conference,times}
\input{math_commands.tex}

\usepackage{hyperref}
\usepackage{url}
\usepackage{graphicx}
\usepackage{booktabs}
\usepackage{pifont}
\usepackage{subcaption}
\usepackage{xcolor}
\usepackage{enumitem}
\usepackage{colortbl}

\iclrfinalcopy
\ificlrfinal
  \newcommand{\liftcode}{\url{https://github.com/AI21Labs/lift-bench}}
  \newcommand{\liftdata}{\url{https://huggingface.co/datasets/ai21labs/LIFTBench}}
\else
  \newcommand{\liftcode}{[anonymized]}
  \newcommand{\liftdata}{[anonymized]}
\fi

\newcommand{\benchmark}{LIFT-Bench}
\newcommand{\benchmarkTrain}{LIFT-Train}
\newcommand{\benchmarkSize}{784}
\newcommand{\benchmarkTrainSize}{34,258}

\title{LIFT: Scoped Constraints for Evaluating and Training Instruction Following}
\author{Eran Goldstein\thanks{Corresponding author: \texttt{erango@ai21.com}} \quad
Yuval Belfer \quad
Yonatan Belinkov \quad
Daniel Gissin \\[0.5em]
\normalfont\small AI21 Labs}

\begin{document}
\maketitle
\lhead{Preprint. Under review.}
\begin{abstract}
\input{sections/abstract.tex}
\end{abstract}
\ificlrfinal

\begin{center}
\begin{tabular}{@{}r@{\ }l@{}}
\textbf{Code:} & \liftcode \\
\textbf{Data:} & \liftdata
\end{tabular}
\end{center}
\fi

\input{sections/introduction.tex}
\input{sections/related_work.tex}
\input{sections/draftingbench.tex}

\input{sections/benchmark_analysis.tex}

\input{sections/draftingtrain.tex}

\input{sections/training_experiments.tex}
\input{sections/conclusion.tex}

\bibliography{iclr2027_conference}
\bibliographystyle{iclr2027_conference}

\newpage
\appendix
\input{sections/appendices.tex}
\end{document}

%% file: math_commands.tex
\usepackage{amsmath,amsfonts,bm}

\def\eqref#1{equation~\ref{#1}}

\def\1{\bm{1}}

\DeclareMathAlphabet{\mathsfit}{\encodingdefault}{\sfdefault}{m}{sl}
\SetMathAlphabet{\mathsfit}{bold}{\encodingdefault}{\sfdefault}{bx}{n}



%% file: sections/abstract.tex
\vspace{-3pt}
Common instruction following (IF) benchmarks, such as IFEval, evaluate the ability of language models (LMs) to follow simple constraints that can easily be verified programmatically. These same constraints, in turn, are used as reward signals for reinforcement learning. This design may lead to reward hacking of constraints, pushing models away from the main task. Recent attempts to address this resulted in benchmarks with richer constraints, but required expert labor, remained closed-access, and cannot be easily extended.
We present \benchmark, an IF benchmark that asks a model to draft a document given a topic and a structured outline.
Constraints are scoped to certain document sections, providing finer control over constraint density.
\benchmark{} includes both verifiable and judged constraints, as well as a training counterpart, \benchmarkTrain.
Importantly, \benchmark{} provides both main task and instruction-following constraints, mitigating shallow reward hacking.
On experiments with multiple models, we find that frontier models struggle with the harder subsets of our benchmark and analyze model performance along multiple axes. We also find that training on \benchmarkTrain{} is much less susceptible to reward hacking than training on the common IFTrain.
\benchmark{} and \benchmarkTrain{} are publicly available, as well as the data generating pipeline.

%% file: sections/introduction.tex
\vspace{-7pt}
\section{Introduction}
\label{sec:introduction}
\vspace{-5pt}

Instruction following (IF) is the ability to satisfy explicit constraints on a
response while still performing the task for which the response is intended. It
is typically measured by benchmarks that programmatically check constraint satisfaction in the language model (LM) response \citep{zhou2023ifeval, pyatkin2025ifbench}.
Early work trained for IF with supervised fine-tuning  \citep{wei2022finetuned,sanh2022multitask,mishra-etal-2022-cross} or reinforcement learning from human feedback \citep[RLHF;][]{NEURIPS2022_b1efde53}.  More recent
efforts train LMs to follow instructions by reusing the same programmatic checkers as RL rewards
\citep{pyatkin2025ifbench}, a kind of reinforcement learning from verifiable rewards \citep[RLVR;][]{lambert2025tulu}.

We highlight four shortcomings of existing IF benchmarks.
First, the specified constraints, and their checkers, are independent of the main task, leading to potential reward hacking \citep{amodei2016concrete,singhal2024a,wang2026reward}.
Second, constraint diversity in  dominant benchmarks is low,
    largely consisting of lexical and formatting constraints \citep{zhou2023ifeval,pyatkin2025ifbench}.
Third,
constraints in many benchmarks are written by humans
    \citep{zhou2023ifeval, wen2024complexbench, he2026advancedif}, which limits size, reproducibility, and adaptation to new domains.
    In addition, the richest sets are also
    eval-only or closed \citep{wen2024complexbench,he2026advancedif}.
Fourth, difficulty is assessed by total constraint count,
ignoring the spread or density of constraints along the response.

We introduce Localized Instruction-Following Tasks (\benchmark), a benchmark of synthetic document-drafting tasks that aims to mitigate these shortcomings. Each \benchmark{} task provides a topic and an outline---a nested tree
of headers whose leaves hold paragraphs, code blocks, tables, and lists---and
attaches constraints to individual elements. Each constraint is  \emph{scoped}: it is evaluated only against the element it is attached to, not
against the whole document.
This distinguishes ``a paragraph in French
somewhere'' from ``section 2.1 in French''.
Scoping allows us to define \emph{density}, how many constraints attach to each section.
This structured design also renders evaluation  more feasible
compared to verifying each constraint against the entire document, as in existing benchmarks.

Constraints in \benchmark{} are either  \emph{verifiable}, using
 deterministic checkers, or \emph{judged},   scored by an LM
judge against generated criteria; the latter cover the content a section must contain.
Together with scoping, this addresses the  above shortcomings:
judged content constraints put main-task adherence inside the reward and extend the vocabulary past lexical and formatting checks; synthesis removes the annotator ceiling
on size and domain; and density becomes a generation parameter rather than a property
the benchmark cannot express.

\benchmark{} contains \benchmarkSize{} examples organized along 10 splits factored by document size, constraint density, and scoping. We also release \benchmarkTrain{}, a corresponding dataset for IF training, consisting of \benchmarkTrainSize{} examples.
We evaluated a range of frontier and open models on \benchmark{}, revealing that models are distinguished by how sharply their scores degrade with increasing density of constraints.
 The globally-scoped constraints used by current IF benchmark is the hardest regime of all, despite carrying an order of magnitude fewer
constraints.
On the training side, training on
\benchmarkTrain{} leads to large improvements on \benchmark, as expected, while training on the common IFTrain \citep{pyatkin2025ifbench} barely improves on our benchmark and often leads to reward hacking. At the same time, training on \benchmarkTrain{} does not degrade on a more challenging out-of-distribution benchmark, AdvancedIF \citep{he2026advancedif}, unlike IFTrain.

To summarize, our contributions are:
\begin{itemize}[leftmargin=20pt,topsep=0pt,parsep=0pt,itemsep=0pt
]
    \item \benchmark{}: a synthetic IF benchmark of document-drafting tasks
    with scoped constraints, combining verifiable and judged scoring in a
    single weighted metric.

    \item \benchmarkTrain{}: the training-side release, which is more robust to reward hacking found when training on lexical-only IF rewards.
    \item An open-source pipeline for synthesizing drafting tasks in
    \texttt{verifiers}-compatible format \citep{primeintellect2025verifiers},
    ready for IF-RLVR training and evaluation.

    \item Analyses that our factored design makes possible: analyses of failures by per-section density and scoping, which require scoped constraints, alongside constraint load, type, and scoring mechanism.
\end{itemize}

%% file: sections/related_work.tex
\vspace{-7pt}
\section{Related Work}
\label{sec:related-work}
\vspace{-5pt}

IFEval~\citep{zhou2023ifeval} was the first benchmark to enable measuring IF, with 541 prompts carrying one to three programmatically checkable
constraints.
IFBench~\citep{pyatkin2025ifbench} counters overfitting to IFEval with held-out
constraint types and publishes IFTrain, a training set that reuses the
checkers as rewards. Both benchmarks verify every constraint against the model's entire
response (global scope), and in both the main task is not reflected in the metric, leading to reward hacking during RL, as we show later.
\citet{pyatkin2025ifbench} also observed such degradation and tried to mitigate it:
completions that already satisfy their constraints are rewarded according to  reward-model-based classifier.
This mechanism tests a response's quality in general, without distinguishing main-task adherence from constraint adherence.
 \benchmark{} aims to mitigate these shortcoming.

A second line of work grades responses against LM-judged checklists. InFoBench
\citep{qin2024infobench} decomposes 500 instructions into 2,250 yes/no
requirements, including content requirements; CFBench \citep{zhang2025cfbench}
scores 1,000 Chinese prompts against prioritized checklists spanning ten
constraint categories; ComplexBench~\citep{wen2024complexbench} scores 1{,}150 human-authored
instructions with
yes/no questions (about 4 constraints per instruction),
using a mix of rules and GPT-4~\citep{openai2023gpt4}.
FollowBench~\citep{jiang2024followbench} grades difficulty by
stacking one constraint per level. These measure richer behavior than the
verifiable benchmarks, but are hand-authored and frozen: no training set and no
synthesis code, limiting flexibility and modularity.

\looseness=-1
AdvancedIF
\citep{he2026advancedif} (1,645 prompts, 6--10 criteria on average) and
ComplexConstraints \citep{mehta2026complexconstraints} (75 public prompts,
10--40 criteria each) use expert-written rubrics and show that rubric rewards
transfer to held-out IF benchmarks. Both depend on expert authoring, which limits
scale and adaptation to new domains, and neither releases its training data;
AdvancedIF also keeps its rubric generator, verifier, and RL code closed.

MOSAIC \citep{purpura2026deconstructing} synthesizes 4,000 prompts with up to
20 constraints each. It identifies the important gap between constraint compliance and task
success, but attempts to resolve it by modularizing the constraints. Their constraints are limited to ``the output's format, style, or structure'', and do not test for content inclusion, as we do. Their constraints are globally scoped, and no training set is released. In \benchmark{}, task adherence is part
of the scoring function itself, and structured tasks scope each constraint to
one section.

As Table~\ref{tab:related-work} summarizes, \benchmark{} is the only IF
benchmark we know of with content constraints, scoped constraints, and a
released training set and pipeline.

\begin{table}[t]
\caption{Comparison with related IF benchmarks.
(\ding{51}): exists but not publicly released.}
\centering
\small
\setlength{\tabcolsep}{4pt}
\begin{tabular}{lrccrcccc}
\toprule
 & \multicolumn{2}{c}{Data} & \multicolumn{4}{c}{Constraints} & \multicolumn{2}{c}{Released} \\
\cmidrule(lr){2-3} \cmidrule(lr){4-7} \cmidrule(lr){8-9}
Benchmark & Size & Authoring & Style & Content & Max & Scope & Train set & Pipeline \\
\midrule
IFEval              & 541   & manual    & \ding{51} & \ding{55} & 3  & global           & \ding{55}   & \ding{55}   \\
IFBench             & 300   & manual    & \ding{51} & \ding{55} & 2  & global           & \ding{51}   & \ding{55}   \\
ComplexBench        & 1,150 & manual    & \ding{51} & \ding{51} & 5  & global           & \ding{55}   & \ding{55}   \\
AdvancedIF          & 1,645 & manual    & \ding{51} & \ding{51} & 10 & global           & (\ding{51}) & (\ding{51}) \\
MOSAIC              & 4,000 & synthetic & \ding{51} & \ding{55} & 20 & global           & \ding{55}   & \ding{51}   \\
ComplexConstraints  & 75    & manual    & \ding{51} & \ding{51} & 40 & global           & (\ding{51}) & \ding{55}   \\
\midrule
\textbf{LIFT-Bench} & 784   & synthetic & \ding{51} & \ding{51} & 80 & section / global & \ding{51}   & \ding{51}   \\
\bottomrule
\end{tabular}
\label{tab:related-work}
\end{table}

%% file: sections/draftingbench.tex
\section{\benchmark}
\label{sec:draftingbench}

\subsection{Task and constraint model}
\label{sec:draftingbench-task}

A \benchmark{} task consists of a topic, an optional outline, a set of constraints, and a scoring configuration (\S\ref{sec:draftingbench-scoring}).
The model is asked to draft a markdown document on the topic; the constraints are what the draft is scored against. See Fig.~\ref{fig:example-task} for an example.

\paragraph{Outline and scope.}
An outline is a nested tree of headers whose leaves hold content elements: paragraphs, code blocks, tables, and lists. Every constraint is attached to one leaf section and is \textbf{scored only against that section's text}, identified by its path in the outline. (A constraint on an artifact---a
table, list, or code block---narrows down, to a single artifact within the
section.)
We call this a \emph{scoped constraint}. Scoping
makes each constraint checkable against a passage rather than  a whole document, allowing for a fine-grained analysis.

\paragraph{Scoping}
We separate tasks to \emph{structured} ones, which come with an outline and scoped constraints, and \emph{flat} ones, which do not have an outline and where
 every constraint scopes globally over the entire response. Flat tasks are the
regime existing IF benchmarks occupy, and we include them for comparison under the same pipeline.

\paragraph{Density.}
Scoping gives an operational definition of \emph{density}: the number of
constraints attached to a single section, the smallest unit of text that the model needs to produce. Density is set at generation time
(\S\ref{sec:draftingbench-pipeline}) rather than inferred after the fact, which
 lets us vary it against a fixed set of documents.\footnote{
 A flat task has one
scope, so its density equals its total constraint count; but that scope is the
whole response rather than one section, so flat and structured densities do not sit on a common scale (\S\ref{sec:draftingbench-splits}).}

\begin{figure}[htbp]
\centering
\fbox{\parbox{\textwidth}{\ttfamily\footnotesize
\raggedright
Write a markdown document about `Ice Hockey World Championships' that has the
following outline: \\[2pt]
\# Ice Hockey World Championships Overview \\
\#\# History and Significance \\
\#\# Format and Qualification Process \\
\#\# Notable Moments and Records \\
\hspace*{2em}[...] \\[2pt]
The `History and Significance' section should follow the constraints below: \\
\hspace*{1em}\textcolor[HTML]{1A4F9C}{*~\parbox[t]{\dimexpr\linewidth-3em\relax}{The text should explore the significance of the championships in
the global ice hockey community, discussing how they contribute to [...]
}} \\
\hspace*{1em}\textcolor[HTML]{1A4F9C}{*~\parbox[t]{\dimexpr\linewidth-3em\relax}{The text should detail the origins of the Ice Hockey World
Championships, tracing back to its first official tournament [...]
}} \\
\hspace*{1em}\textcolor[HTML]{7B2D8E}{*~\parbox[t]{\dimexpr\linewidth-3em\relax}{The text should contain the word `tournament' at least 3 times.}} \\
\hspace*{1em}\textcolor[HTML]{7B2D8E}{*~\parbox[t]{\dimexpr\linewidth-3em\relax}{The text should be written in German.}} \\
The `Format and Qualification Process' section should follow the constraints
below: \\
\hspace*{2em}[...] \\
The `Notable Moments and Records' section should follow the constraints below: \\
\hspace*{2em}[...]
}}
\caption{Example task in \benchmark{}, with  verifiable (\textcolor[HTML]{7B2D8E}{purple}) and judge-based constraints (\textcolor[HTML]{1A4F9C}{blue}).
}
\label{fig:example-task}
\end{figure}

\subsection{Scoring}
\label{sec:draftingbench-scoring}

Every task carries the list of constraints it will be evaluated against, and its
score is a weighted sum of per-constraint results in $\{0,1\}$. Constraints are
grouped into \textbf{structure} (outline compliance) and \textbf{body}
(everything else). The relative weight of the two groups is a configuration
parameter.
In our
released configuration, structure receives 10\% and content 90\%. The same scalar
serves as the benchmark metric and  as the RL reward
(\S\ref{sec:draftingtrain}).

The LM judge we use is Gemma-4-31b~\citep{gemmateam2026gemma4}; see Appendix~\ref{sec:appendix-judge} for more details. We ran a human validation study to evaluate its quality, and found it achieved an F1 score of 0.73 against the human annotations; see Appendix~\ref{sec:appendix-judge-meta} for full details.

Because content constraints carry most of the weight and include per-section
content requirements, a response that satisfies every formatting constraint
without drafting a meaningful document scores close to zero.
This helps avoid reward hacking when training, as we show later (\S\ref{sec:training-experiments}).

\subsection{Synthesis pipeline}
\label{sec:draftingbench-pipeline}

\begin{figure}[htbp]
\centering
\includegraphics[width=\textwidth]{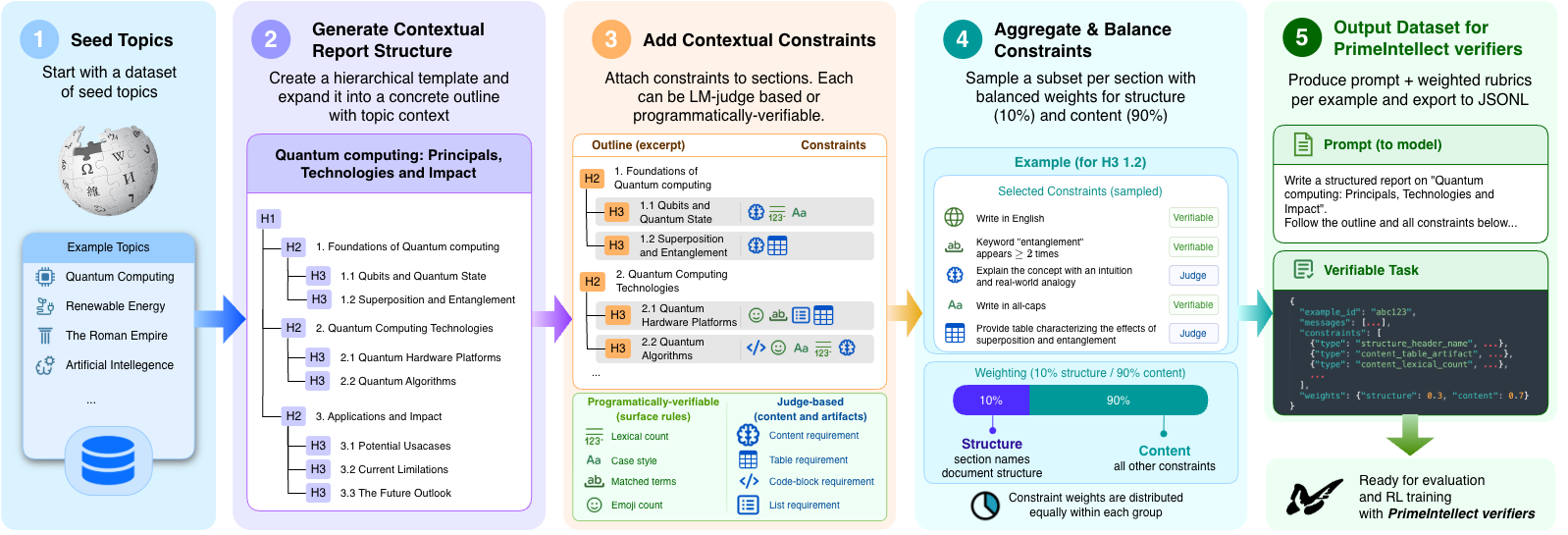}
\caption{Synthesis pipeline overview.}
\label{fig:pipeline}
\end{figure}

The pipeline is a sequence of atomic stages, outlined in  Fig.~\ref{fig:pipeline}:\footnote{ Appendix~\ref{sec:appendix-pipeline} gives the full  description, including prompts, models, and   parameters  for each released split.}
\begin{enumerate}[leftmargin=15pt,itemsep=2pt,topsep=1pt,parsep=1pt]
\item \textbf{Topic seeding}: sample seed topics from Wikipedia.
\item \textbf{Outline construction}: (a)  outline templating, where we generate structural document skeletons.; and (b) outline expansion, where  an LM fills each skeleton into a concrete, topic-specific outline.
\item \textbf{Constraint generation}: parallel per-type generators propose
candidate constraints over the outline, one block per constraint type, covering
both mechanisms.
\item \textbf{Task configuration}
  \begin{enumerate}
  \item Constraint sampling: draw the task's constraints from the candidate pool, with  knobs for setting how many constraints each section carries, how many sections carry constraints at all, and which constraint types appear.
  \item Reward shaping: set the structure/content budget split for the task.
  \end{enumerate}
\item \textbf{Task Scoping}: render the task as a prompt
(Fig.~\ref{fig:example-task}). Scoping can be \textbf{structured}, the
outline with its per-header constraints, or \textbf{flat}, an IFEval-style
prompt in which the same constraints are stated as applying to the entire output.
The flat form allows us to compare scoped and globally-scoped prompting
directly, holding topic, outline, and constraint content fixed
(\S\ref{sec:benchmark-analysis}).
\end{enumerate}

\subsection{The released splits}
\label{sec:draftingbench-splits}

Ten splits come from three generation runs: short outlines, long outlines,
and flat (no outline). Within each run, every split shares the same topics and outlines;
only the number of constraints attached to them changes. Appendix~\ref{sec:splits-stats} provides descriptive statistics.

Splits are comparable only if they share the same scoping (flat or structured). A flat split labelled
\emph{many} carries far fewer constraints in absolute terms than a structured
split with the same label, because a flat task has no sections across which to
distribute them---every constraint lands in a single scope. The ten splits are
therefore not a monotonic difficulty scale.

%% file: sections/benchmark_analysis.tex
\section{Benchmark Evaluation Experiments}
\label{sec:benchmark-analysis}
\subsection{Setup}
\label{sec:analysis-setup}

We evaluate 17 models on the 10 released splits.
Every example is generated 8 times; an example's score is the mean over its 8 runs, and the
intervals are 95\% confidence intervals across examples. Judged constraints are scored by Gemma-4-31b~\citep{gemmateam2026gemma4}.

The nine main ones are \textbf{(1) closed-weight:} GPT-5.5~\citep{openai2026gpt55},
GPT-5.6 Sol~\citep{openai2026gpt56}, GPT-5 nano~\citep{openai2026gpt5card}, Claude Fable
5~\citep{anthropic2026fable5} and Claude Opus 4.8~\citep{anthropic2026opus48}; and
\textbf{(2) open-weight:} DeepSeek (DS) V4-Pro~\citep{deepseek2026v4},
GLM 5.3~\citep{glm2026glm5}, Kimi K3~\citep{kimi2026k3} and
Qwen3.8-2.4T-A95B~\citep{qwen2026qwen38}. Appendix~\ref{sec:appendix-models} lists the full
set of evaluated models.

\subsection{Overall results}
\label{sec:analysis-overall}

Table~\ref{tab:leaderboard} reports per-split scores for a selected 9 models;
Appendix~\ref{sec:appendix-full-results} gives the full results for all 17.
Structured splits are close to ceiling for the strongest models (GPT-5.5 averages 0.96), but the benchmark is not saturated: no model exceeds 0.85 on any flat split, and weaker models fall to 0.47 on the densest structured splits.

\begin{table}[h]
\centering
\caption{Mean score per split; best per row in bold;
averages are $\pm$95\% CI; `avg (all)' weighs the structured and flat averages equally.}
\label{tab:leaderboard}
\resizebox{\textwidth}{!}{%
\input{figures/benchmark_analysis/leaderboard.tex}
}

\end{table}

Kimi K3 and DeepSeek V4 Pro have the two highest overall averages: they match the strongest closed models on structured splits and surpass them on flat ones. The two scopes also reorder the ranking: GPT-5.5 is first on structured splits but fifth on flat\footnote{Constraint count does not explain the drop of closed models; see Appendix~\ref{sec:appendix-flat-anomaly}.}, while Qwen3.8 rises from seventh to third.
As all other benchmarks only contain globally scoped constraints, \benchmark{} provides a novel perspective on model rankings.

\subsection{Where models fail: load, mechanism, and type}
\label{sec:analysis-fine-grained}

\begin{figure}[htbp]
    \centering
    \includegraphics[width=0.9\textwidth]{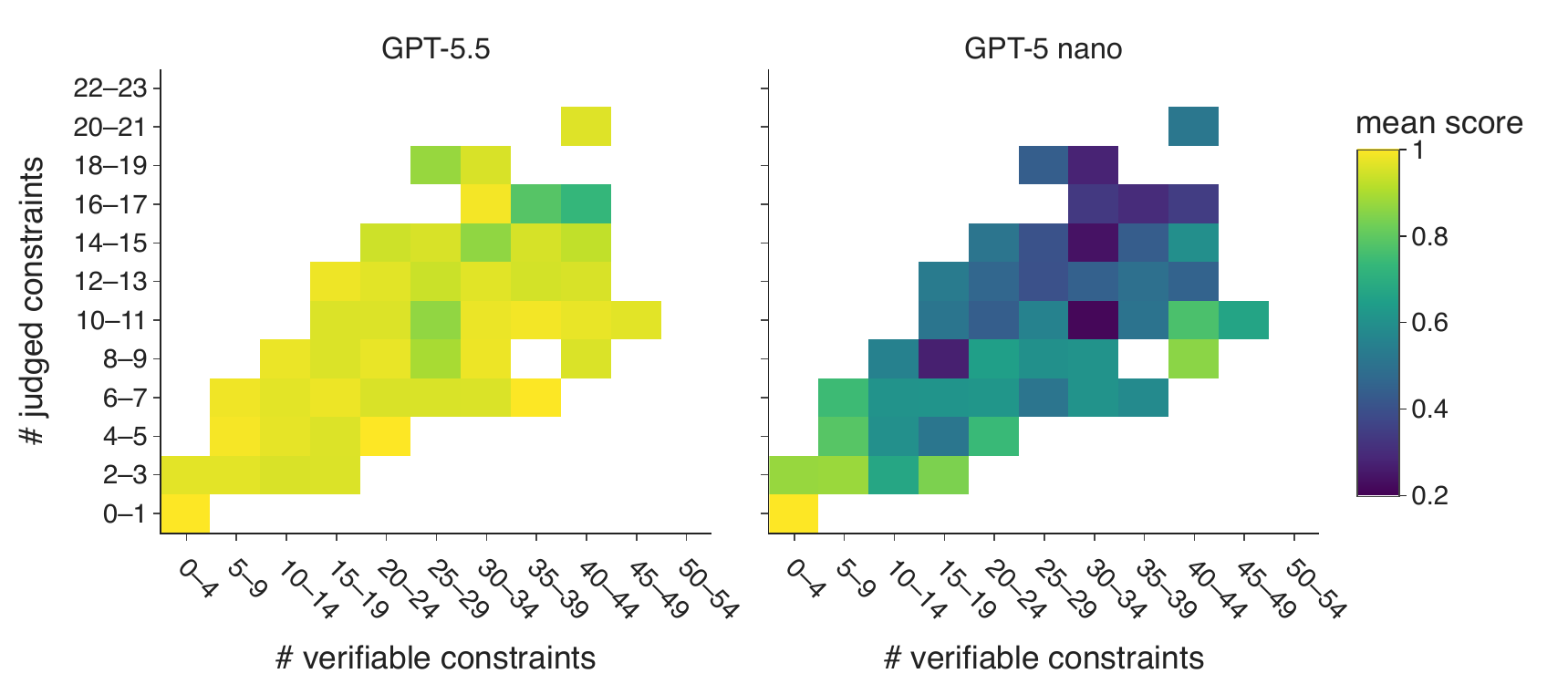}
   \caption{Benchmark scores broken by number of judged and verifiable constraints.
   In the
lightest cell (bottom left) GPT-5.5 and GPT-5 nano both score near 1, but at the
densest loads the weaker model falls to 0.2--0.5 while the stronger stays above
0.7.
}
    \label{fig:load-sensitivity-pair}
\end{figure}

Intuitively, the more constraints are added, the more difficult it is to follow the instruction.
To investigate this, we compare two models: the top frontier model GPT-5.5 and the weaker GPT 5 nano.
We break down performance by the number of judged versus verified constraints. \benchmark{} uniquely supports this analysis
thanks to its factored design.
As Fig.~\ref{fig:load-sensitivity-pair} shows, GPT-5.5 declines mildly across the entire plane, while the weaker nano model collapses, and does so more steeply along the judged axis.
In other words, the number of judged constraints affects performance more than the number of verifiable constraints, in weaker models.

Next, we break down the results by constraint density, that is, the number of constraints per section. To do so, we fix a section density and increase the total load, by adding constraints elsewhere in the document.
As Table~\ref{tab:section-load} shows, frontier models are only mildly affected by this
intervention (from near zero for Kimi K3 to $-1.5\%$ per density step for GPT-5.5), while
smaller ones degrade steadily ($-5.5\%$ for GPT-5 nano), indicating that their ability to satisfy a
given local constraint is affected by more remote constraints. The results over all evaluated models are given in
Appendix~\ref{sec:appendix-full-heatmaps}.\footnote{Note that the globally-scoped regime is a different case. There, a flat task means that every constraints belongs in a single global scope, which the LM must satisfy. This explains why the frontier models' robustness to the rest of the document does not help them on flat tasks.}

\begin{table}[htbp]
\centering
\caption{
Change in section score, estimated via linear regression and averaged over density types.
At fixed per-section density, frontier models are only mildly affected by constraints
elsewhere in the document while smaller models continue to degrade.
The full results are in Table~\ref{tab:section-load-all} and the fitting procedure in
Appendix~\ref{sec:appendix-section-load}.}
\label{tab:section-load}
\input{figures/benchmark_analysis/section_load_slopes.tex}
\end{table}

\begin{figure}[htbp]
\centering
\includegraphics[width=0.9\textwidth]{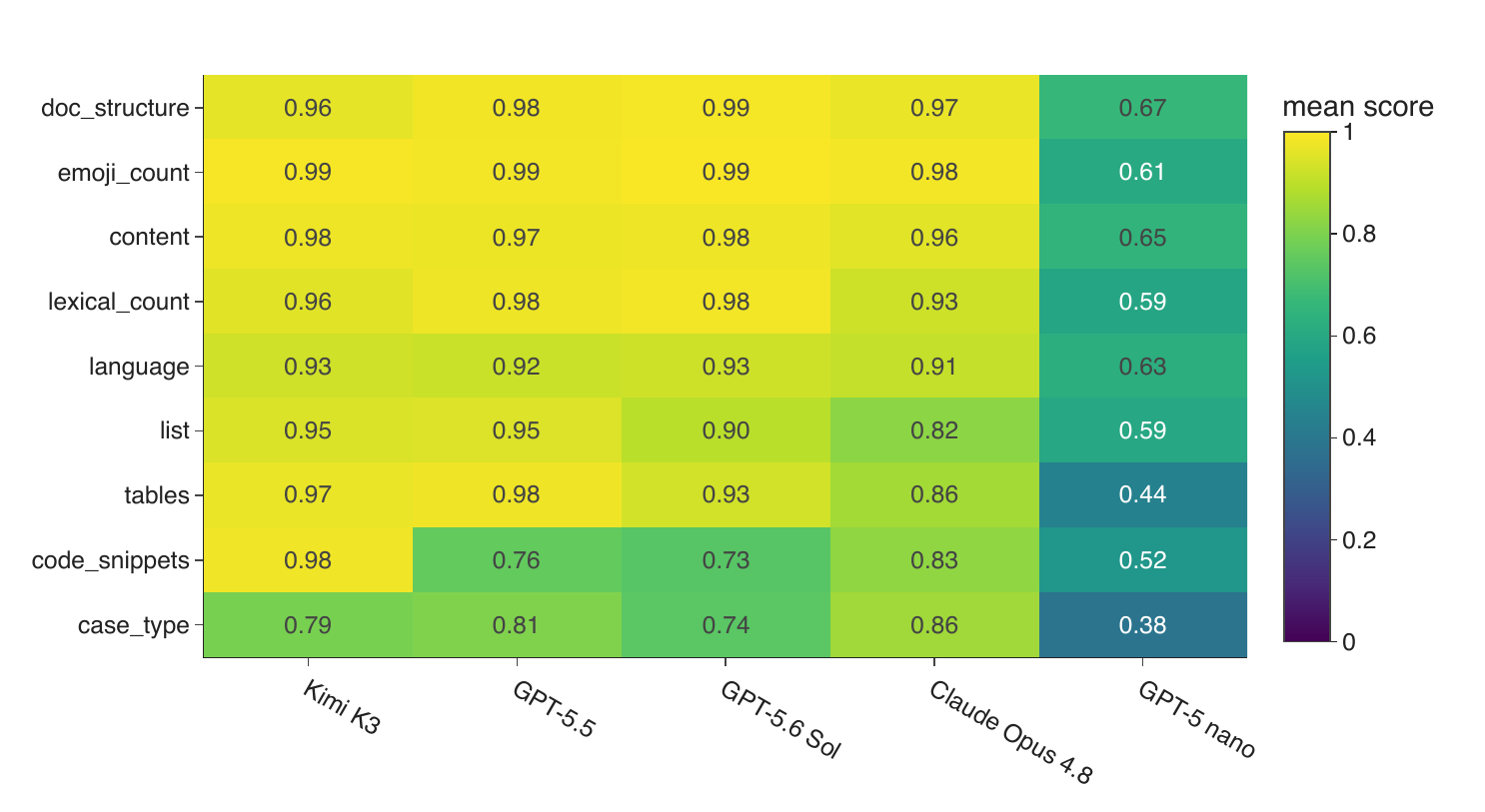}
\vspace{-3pt}
\caption{Benchmark scores per constraint type.
Models display distinct constraint-type fingerprints, even where their averages are close.
}
\label{fig:type-fingerprint}
\vspace{-5pt}
\end{figure}

Finally, we analyze how models follow different types of instructions
(Fig.~\ref{fig:type-fingerprint}).
Some types are near ceiling for every frontier model (\texttt{doc\_structure}, \texttt{emoji\_count}, \texttt{content}), while artifact constraints (\texttt{list}, \texttt{tables}, \texttt{code\_snippets}) and \texttt{case\_type} vary most. GPT-5 nano trails the frontier models on every type, most on
\texttt{tables} and \texttt{case\_type}; \texttt{list} and \texttt{tables} are where mid-tier models such as Claude Sonnet 4.6 fall behind
(Appendix~\ref{sec:appendix-full-heatmaps}).

The frontier models agree on most types but not all: on \texttt{code\_snippets} Kimi K3
scores 0.98 against 0.73--0.83 for the other three. What the breakdown shows is a shape rather than a ranking: models at a similar average score can fail at different things. Full grids are in
Appendix~\ref{sec:appendix-full-heatmaps}.

Load and type breakdowns are also possible on MOSAIC~\citep{purpura2026deconstructing}. Per-section density and scoping are not, because existing IF benchmarks check every constraint against the whole response; in \benchmark{} both are generation parameters rather than incidental properties of the prompts.

%% file: figures/benchmark_analysis/leaderboard.tex
\begin{tabular}{lccccccccc}
\toprule
 & \multicolumn{3}{c}{GPT} & \multicolumn{2}{c}{Claude} & \multicolumn{4}{c}{Open Models} \\
\cmidrule(lr){2-4} \cmidrule(lr){5-6} \cmidrule(lr){7-10}
split & 5.5 & 5.6 Sol & 5 nano & Fable 5 & Opus 4.8 & DS V4 Pro & GLM-5.3 & Kimi K3 & Qwen3.8-A95B \\
\midrule
short/few & \textbf{0.97} & 0.95 & 0.71 & 0.92 & 0.94 & 0.95 & 0.87 & 0.97 & 0.89 \\
short/some & \textbf{0.96} & 0.94 & 0.62 & 0.91 & 0.92 & 0.95 & 0.88 & 0.96 & 0.88 \\
short/many & \textbf{0.98} & 0.94 & 0.58 & 0.82 & 0.89 & 0.93 & 0.83 & 0.97 & 0.82 \\
short/numerous & \textbf{0.93} & 0.92 & 0.47 & 0.81 & 0.82 & 0.92 & 0.80 & 0.92 & 0.81 \\
long/few & \textbf{0.98} & 0.96 & 0.67 & 0.92 & 0.96 & 0.95 & 0.93 & 0.96 & 0.90 \\
long/some & \textbf{0.95} & 0.95 & 0.61 & 0.91 & 0.91 & 0.94 & 0.89 & 0.95 & 0.89 \\
long/many & \textbf{0.97} & 0.93 & 0.55 & 0.90 & 0.92 & 0.95 & 0.87 & 0.95 & 0.87 \\
long/numerous & 0.91 & 0.89 & 0.47 & 0.83 & 0.87 & 0.90 & 0.84 & \textbf{0.92} & 0.80 \\
flat/some & 0.71 & 0.66 & 0.73 & 0.64 & 0.71 & 0.82 & 0.60 & \textbf{0.85} & 0.71 \\
flat/many & 0.72 & 0.65 & 0.68 & 0.66 & 0.74 & 0.80 & 0.60 & \textbf{0.83} & 0.76 \\
\midrule
avg (structured) & \textbf{0.96}{\scriptsize$\pm$0.01} & 0.94{\scriptsize$\pm$0.01} & 0.59{\scriptsize$\pm$0.03} & 0.88{\scriptsize$\pm$0.01} & 0.90{\scriptsize$\pm$0.01} & 0.94{\scriptsize$\pm$0.01} & 0.86{\scriptsize$\pm$0.01} & 0.95{\scriptsize$\pm$0.01} & 0.86{\scriptsize$\pm$0.01} \\
avg (flat) & 0.71{\scriptsize$\pm$0.03} & 0.66{\scriptsize$\pm$0.03} & 0.71{\scriptsize$\pm$0.04} & 0.65{\scriptsize$\pm$0.03} & 0.72{\scriptsize$\pm$0.03} & 0.81{\scriptsize$\pm$0.03} & 0.60{\scriptsize$\pm$0.03} & \textbf{0.84}{\scriptsize$\pm$0.03} & 0.73{\scriptsize$\pm$0.03} \\
avg (all) & 0.84{\scriptsize$\pm$0.02} & 0.80{\scriptsize$\pm$0.01} & 0.65{\scriptsize$\pm$0.02} & 0.76{\scriptsize$\pm$0.02} & 0.81{\scriptsize$\pm$0.02} & 0.87{\scriptsize$\pm$0.02} & 0.73{\scriptsize$\pm$0.01} & \textbf{0.90}{\scriptsize$\pm$0.02} & 0.79{\scriptsize$\pm$0.02} \\
\bottomrule
\end{tabular}

%% file: figures/benchmark_analysis/section_load_slopes.tex
\begin{tabular}{lrrr}
\toprule
 & \multicolumn{2}{c}{\% score change by document size} & \\
\cmidrule(lr){2-3}
Model & short & long & mean \\
\midrule
Kimi K3 & $-0.9$\% & $+0.2$\% & $-0.4$\% \\
Claude Opus 4.8 & $-0.4$\% & $-0.5$\% & $-0.4$\% \\
GPT-5.6 Sol & $-0.4$\% & $-0.9$\% & $-0.7$\% \\
GPT-5.5 & $-1.3$\% & $-1.7$\% & $-1.5$\% \\
GPT-5 nano & $-7.0$\% & $-4.1$\% & $-5.5$\% \\
\bottomrule
\end{tabular}

%% file: sections/draftingtrain.tex
\section{\benchmarkTrain}
\label{sec:draftingtrain}

\benchmarkTrain{} is the training-side release of the same pipeline: training and
test tasks are produced by one generator, and the benchmark's scoring function
(\S\ref{sec:draftingbench-scoring}) is used unchanged as the RL reward. Accordingly, most of the reward consists of content constraints, so an output that satisfies every formatting
constraint without drafting a meaningful document earns almost nothing.

In contrast, in IFTrain \citep{pyatkin2025ifbench} constraints are lexical and verifiable but do not test whether the response's content meets the request.
\citet{pyatkin2025ifbench} acknowledge the potentially-hackable reward and add a general-purpose classifier, based
on a reward model, on top of the verifiable reward, bonusing or penalizing rollouts by the classifier's output. However, that signal rates overall quality without distinguishing satisfaction of the verifiable constraints from adherence to the requested task. In \benchmarkTrain{},
task adherence is scored by constraints of its own, inside the same reward, rather than by a separate quality signal.
Thus, we hypothesize that \benchmarkTrain{} is less likely to be prone to reward hacking.
We perform multiple training runs on various configurations from \benchmarkTrain{}, as well as comparing with IFTrain, to isolate which aspects are more important.

\vspace{-3pt}
\subsection{Training mixtures and setup}
\label{sec:draftingtrain-mixtures}
\vspace{-3pt}

\paragraph{Training mixtures.}
Using the factored splits of \S\ref{sec:draftingbench-splits}, we vary the
training mixture along two independent axes: training  data composition and curriculum.

\vspace{-2pt}
\begin{description}[itemsep=1pt,parsep=0pt,topsep=0pt]
    \item[Composition] We consider four data compositions: \emph{Structured-only} only contrains structured data splits;
\emph{Flat-only} only contains flat splits  (closest to IFEval-style data);
 \emph{Combined} contains both structured and flat data (controlled for size); and
 \emph{Combined + IFBench constraints} adds IFBench-style
constraint types to the pool.
\item[Curriculum] Each composition is trained either at \emph{fixed difficulty},
sampling across splits without regard to load, or on a \emph{curriculum} that presents splits from easy to hard.
\end{description}

As a control we train on IFTrain (lexical, verifiable-only
constraints). The flat-only \benchmarkTrain{}
mixture is the closer comparison: it matches IFTrain's prompt format and
constraint style, and differs in that its reward includes a content component.%
\footnote{%
One caveat for this comparison is that prompt content still differs: our topics are wiki-seeded drafting tasks, while IFTrain's are taken from WildChat~\citep{zhao2024wildchat}.}

\paragraph{Training setup.}
We use GRPO~\citep{shao2024deepseekmath} on three base models: Qwen2.5 7B and 32B (non-reasoning)~\citep{qwen2024qwen25} and Qwen3-32B (reasoning)~\citep{yang2025qwen3} for a total of 180 steps; the Qwen2.5 models continue to step 390 to look for long-training pathologies.

%% file: sections/training_experiments.tex
\section{Training Experiments}
\label{sec:training-experiments}

\subsection{Main training results}

Table \ref{tab:training_recipe_deltas} compares the results of training on our \benchmarkTrain{} with training on the common IFTrain \citep{pyatkin2025ifbench}.
As expected, training on IFTrain performs well on IFEval and IFBench, as IFTrain was designed to mimic their constraints. Similarly, training on \benchmarkTrain{} performs well on \benchmark{}.
However, considering more out-of-distribution benchmarks, AdvancedIF and ComplexBench (\S\ref{sec:related-work}), training on IFTrain degrades performance while \benchmarkTrain{} provides small improvements. This suggests that  IFTrain  leads to more reward hacking than \benchmarkTrain{}.

\input{figures/training_experiments/delta_table}

To probe this, we examined the responses by Qwen2.5-7B over the course of training on IFTrain.
By step 180, the reward saturates at $1.0$ while median visible length on the held-out validation
prompts plummets from 114 to 28.5 words, and to 18 by step 390 (Fig.~\ref{fig:reward-hacking-length}).
Some of the cases we observed in the GRPO rollouts demonstrate reward hacking: by step 180, bare constraint tokens with high rewards; by step 390 we see empty section headers, long repetitions of a single phrase, and other pathological behaviors. See Appendix \ref{app:reward-hacking} for some example rollouts as well as more quantitative analysis.
In contrast, the same base model trained on \benchmarkTrain{} does not exhibit this high reward-hacking tendency during training.

We attribute this to the content-adherence component in \benchmarkTrain{}.
However,
one potential confound with this comparison is that our benchmark differs from IFTrain in more than the reward structure (different format, structure, and topics).
To control for this, we compared also with training on a flat-only version of \benchmarkTrain{}. It matches IFTrain's surface form (flat prompts, IFEval-style constraints), but its reward includes  content adherence. We found this set to produce healthy training, without length pathologies as those mentioned above.
While the comparison to IFTrain is not perfect (prompts still cover different topics), this is another evidence that content adherence mitigates reward hacking.

\subsection{Mixture composition and training order}
\label{sec:training-mixtures}

\begin{figure}[h]
    \centering
    \includegraphics[width=\textwidth]{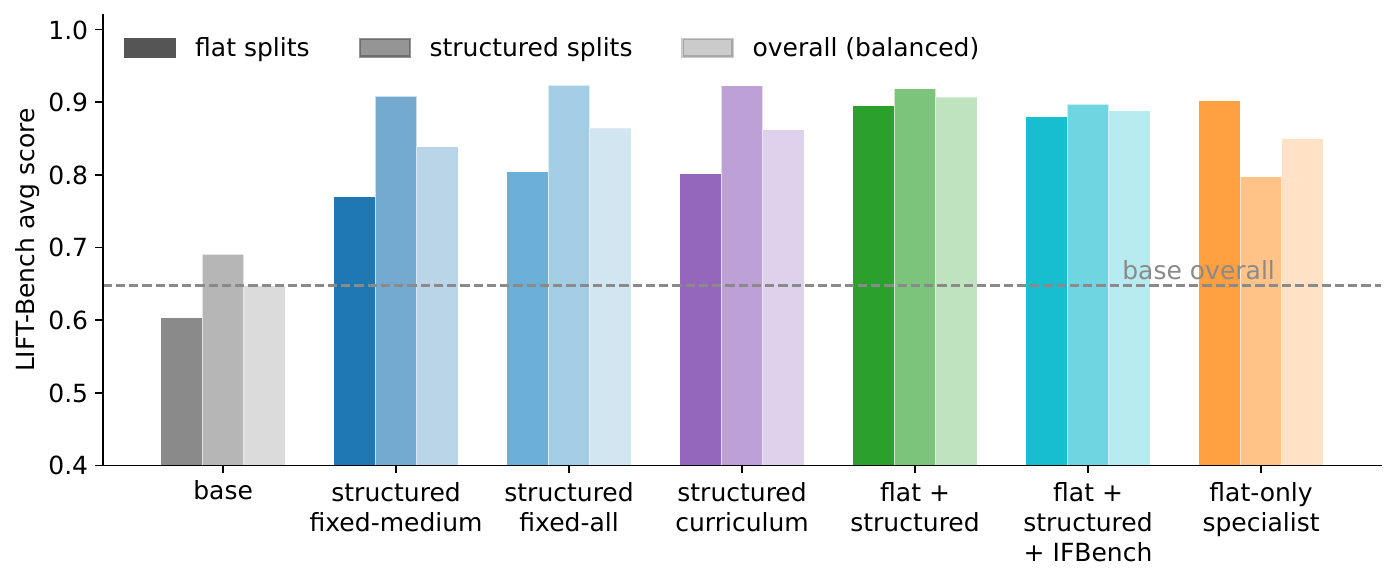}
    \caption{Comparison of training mixtures on Qwen3-32B: \benchmark{} score at GRPO
    step 180, averaged over the flat splits, the structured splits, and their macro-average
    (mean of the flat and structured averages).}
    \label{fig:training-mixtures}
\end{figure}

Next, we examine the role of the training set composition and curriculum, training
Qwen3-32B on each mixture (Fig.~\ref{fig:training-mixtures}).
The mixtures seem largely interchangeable when inspecting the overall scores. However, when
breaking down performance to different splits in \benchmark{}, we do see some composition
effects: training on a single scoping also improves the other, but falls short of the
combined mixture, which does well on both: structured-only training (fixed-all) reaches
0.805 on the flat splits against 0.896 for the combined mixture, and flat-only training
reaches 0.798 on the structured splits against 0.919 (base: 0.604 flat, 0.691 structured).

Training curriculum of easy-to-hard splits also makes no difference: the structured
curriculum run scores on par with its fixed-difficulty counterpart
(Fig.~\ref{fig:training-mixtures}). This may be due to the split design
(\S\ref{sec:draftingbench-splits}): adjacent splits overlap, so ordering does not add that
much diversity. We leave deeper exploration of training curriculum to future work.

%% file: figures/training_experiments/delta_table.tex
\providecommand{\cellcolor}[2][]{}
\providecommand{\benchmark}{LIFT-Bench}
\providecommand{\benchmarkTrain}{LIFT-Train}
\definecolor{deltapos}{HTML}{1B7F3B}
\definecolor{deltaneg}{HTML}{C0392B}
\definecolor{indist}{HTML}{DFE6EC}
\begin{table}[htbp]
\centering
\caption{Benchmark scores after GRPO training on Qwen2.5-7B (step 180), with the change in points against the base model in parentheses. Shaded cells are in-distribution;  other cells are out-of-distribution. Each recipe gains most on the benchmark its reward can see, but only \benchmarkTrain{} also improves the external judged benchmarks.}
\label{tab:training_recipe_deltas}
\begin{tabular}{lrcc}
\toprule
\textbf{Benchmark} & \textbf{base} & \textbf{IFTrain} & \textbf{\benchmarkTrain{}} \\
\midrule
IFEval & 0.712 & 0.764 {\scriptsize\textcolor{deltapos}{($\mathbf{+5.2}$)}} & 0.723 {\scriptsize\textcolor{deltapos}{($+1.1$)}} \\
IFBench & 0.270 & \cellcolor{indist}0.430 {\scriptsize\textcolor{deltapos}{($\mathbf{+16.0}$)}} & 0.278 {\scriptsize\textcolor{deltapos}{($+0.8$)}} \\
\benchmark{} (10-split avg) & 0.423 & 0.451 {\scriptsize\textcolor{deltapos}{($+2.8$)}} & \cellcolor{indist}0.822 {\scriptsize\textcolor{deltapos}{($\mathbf{+39.9}$)}} \\
AdvancedIF (3-split avg) & 0.392 & 0.374 {\scriptsize\textcolor{deltaneg}{($-1.8$)}} & 0.401 {\scriptsize\textcolor{deltapos}{($+0.9$)}} \\
ComplexBench (EN) & 0.324 & 0.316 {\scriptsize\textcolor{deltaneg}{($-0.8$)}} & 0.327 {\scriptsize\textcolor{deltapos}{($+0.3$)}} \\
\bottomrule
\end{tabular}
\end{table}

%% file: sections/conclusion.tex
\section{Conclusion}
\label{sec:conclusion}

Instruction-following is measured by checking constraints and trained by
rewarding the same checks, and both halves of that arrangement are blind to whether the requested task was performed. \mbox{\benchmark{}} closes the blindspot by scoping constraints to individual sections and scoring content requirements alongside formatting ones, which makes constraint density a controlled variable and main-task adherence part of the reward rather than an assumption about it.
Our experiments with various closed frontier models and leading open models demonstrate that \benchmark{} is a useful benchmark for evaluating instruction following. Its modular composition exposes weaknesses of different models along constraint density, mechanism, and type. In addition,  the released training counterpart, \benchmarkTrain{}, is less susceptible to reward hacking than the common IFTrain, thanks to the content-adhering part of its reward.

There are a few possible extensions for \benchmark{} to pursue. A finetuned rubric judge and a learned rubric generator, in the style of AdvancedIF's verifier \citep{he2026advancedif}, would harden the scoring stage that the benchmark and the reward both rely on, and can be easily added to the pipeline. The task distribution can grow along further axes: richer constraint compositions, adding the conditional and chained dependencies of ComplexBench \citep{wen2024complexbench} to our conjunctive ones, and multi-turn splits, where an editorial follow-up tests ongoing edits rather than drafting alone.


\benchmark{}, \benchmarkTrain{} and the synthesis pipeline are available at \liftcode, and the datasets at \liftdata.

%% file: sections/appendices.tex
\input{appendix/data_pipeline.tex}
\input{appendix/judge_scoring.tex}
\input{appendix/judge_meta_eval.tex}

\input{appendix/splits}

\input{appendix/models.tex}

\input{appendix/full_results.tex}
\input{appendix/flat_anomaly.tex}

\input{appendix/full_heatmaps.tex}

\input{appendix/section_load_by_setting.tex}

\input{appendix/reward_hacking_analysis}

%% file: appendix/data_pipeline.tex
\section{Synthesis Pipeline: Detailed Description}
\label{sec:appendix-pipeline}

This appendix expands the pipeline sketch of \S\ref{sec:draftingbench-pipeline}. Each
paragraph below corresponds to one stage of Fig.~\ref{fig:pipeline}, and reports the
concrete parameter values used to synthesise the ten released splits
(\S\ref{sec:draftingbench-splits}). Values that are shared by every released
configuration are given inline; values that vary across splits are summarised in
Table~\ref{tab:pipeline-splits}, and the remaining fixed hyper-parameters in
Table~\ref{tab:pipeline-fixed}.

\paragraph{Implementation and execution model.}
The pipeline is a chain of blocks. A block consumes the output of its
predecessor and
writes its own output to a dedicated directory. The pipeline implements a per-block caching mechanism, so a sweep that changes only a late stage reuses every earlier artifact. This
means that all splits that share a document-size
setting also share  the same outlines and the same pool of candidate
constraints, and differ only in how that pool is sampled and verbalized.
The
density axis is therefore a controlled comparison over identical documents. Items that fail
at any stage (e.g., an LLM response with no parseable code block, or an outline that does not match its skeleton)  are annotated with an \texttt{error} field, diverted to a per-block
error file and dropped from the stream; a block aborts the run if fewer than $20\%$ of its items survive.  (\texttt{min\_success\_threshold} $=0.2$).
Every LLM call in the pipeline (outline expansion, constraint proposal and keyword generation) is served by the
same off-the-shelf model, Qwen2.5-32B-Instruct, at temperature $0.6$; the constraint
samplers use seeded randomness for reproducibility.

\paragraph{Topic seeding.}
Seed topics are the core of the generated drafting task. We used a list of article titles of the Wikipedia \emph{Vital Articles} list, level
5~\citep{wikipedia2023vital}. We work from a historical snapshot of that list, containing approximately $40{,}000$ unique topics.
Titles are deduplicated and truncated to the first $4000$. Each topic is mapped to a unique identifier \texttt{example\_id} that will be used for train/test splitting.

In general, the topic seed is not restricted to encyclopedic subjects.
It is consumed only as free text at outline expansion and as context for judge-based metrics, and never matched
against a knowledge base, so any thing one might write a blog post or an article about is an admissible topic: open-ended
questions (``Why are video games so addictive?''), overview titles (``Gaming
consoles: an overview''), practical subjects (``Useful cooking
skills''), etc. . \emph{Wiki Vital Articles} is a large and broadly-covering list, and it can be swapped with
any topic source without touching the rest of the pipeline.

\paragraph{Outline templating.}
Before any topic-specific text is generated, each item is assigned a pseudo-random
\emph{skeleton}: a tree with a maximum depth $D$ and a maximum number of children per tree-node $C$.
The tree corresponds to the outline's structure: each node corresponds to a section, and they are structured according to the tree's hierarchy.
We use two structured configurations:
 \emph{short document} ($D=2$ and $C=3$) and  \emph{long document} ($D=2$ and $C=5$).
\emph{short} documents yield $5$ headers and $4$ leaf
sections per document on average, and \emph{long} ones yield $11$ headers and $9$ leaves on average.

We also define \emph{Flat} tasks, setting both parameters to $0$, which degenerates to a single H1 node, the single scope into which every constraint is targeting.

\paragraph{Outline expansion.}
The skeleton and the topic are handed to the LLM, which is asked to return an outline with that \emph{exact} skeleton.
We validate the model's response to make sure the response has the exact skeleton structure we requested, with no additional sections/subsections. Invalid completions are dropped from the pipeline. In our setting it was a rare issue ($<1\%$ of items).

That way we obtain a plausible outline for a document about our selected topic.

We use Qwen-2.5-32b-Instruct for that task, with $\le 2048$ tokens for the completion.

\paragraph{Constraint generation.}
The next stages populate the outline with constraints. A \emph{constraint} is a criterion
that a text must satisfy, paired with a checker that returns a boolean for it. A constraint
is scoped to a single section. Each constraint type is generated by its own atomic block;
the blocks run in parallel over the outline, attaching candidate constraints to its sections
without reading one another's output. Extending the benchmark with a new constraint type is
largely a matter of adding one block. The two families are distinguished by how their
constraints are checked:verified deterministically, or judged by an LLM critic
(Appendix~\ref{sec:appendix-judge}).

\paragraph{Constraint families.}
Most types propose several candidates per section, which the sampling stage below then draws
from. Thus a block's output is a pool of admissible constraints, not the task's final load.
A few types instead need their candidates chosen with care, to avoid conflicting instructions: a section carries at most one language constraint, and the IFBench constraints are sampled from a curated list under a mutual-compatibility check.
We do not resolve conflicts globally: nothing checks a task's constraints against one another, so avoiding contradictions is left to each type at generation time. This is a
limitation---adding a type means designing it not to collide with the existing ones---but this keeps the blocks independent and the pipeline a simple parallel fan-out.
Table~\ref{tab:constraint-sampling} gives each type and where its candidates come from.

\begin{table}[htbp]
  \centering
  \caption{Constraint types and the source of each block's candidates. Keywords for \texttt{lexical\_count} constraints are LLM-generated to make them relevant to the discussed topic and not arbitrary. We used \texttt{ifbench\_constraint} for ablations, but not in the final splits. Shared
  hyper-parameters are in Table~\ref{tab:pipeline-fixed}.}
  \label{tab:constraint-sampling}
  \footnotesize
  \setlength{\tabcolsep}{4pt}
  \begin{tabular}{lll}
  \toprule
  \textbf{Family} & \textbf{Type} & \textbf{Candidate source} \\
  \midrule
  Judged & \texttt{content} & LLM-based \\
   & \texttt{tables}/\texttt{list}/\texttt{code\_snippets} & LLM-based; often empty if not relevant to the topic \\
  \midrule
  Verifiable & \texttt{language} & sampled over ten fixed languages \\
   & \texttt{case\_type} & sampled over \{upper, lower, title\} \\
   & \texttt{emoji\_count} & sampled over \{in-range, more-than, less-than\} \\
   & \texttt{lexical\_count} & keyword count / must-appear / must-not-appear \\
   & \texttt{ifbench\_constraint} & subset of IFTrain constraints; excluded from released splits \\
  \bottomrule
  \end{tabular}

\end{table}

\paragraph{Constraint sampling.}
Sampling runs once per family: one pass over the verifiable pool, one over the judged
pool. Each pass draws \emph{up to} $N$ constraints per section rather than exactly $N$,
which keeps density varied within a split (Fig.~\ref{fig:splits-identity-card}).
Constraints can be sampled globally, over the whole document at once, or per-section; we
use the latter throughout, since it makes constraint density a quantity we set rather than a
by-product of document length. A pass also restricts itself to at most $M$ leaf sections,
drawn uniformly without replacement, which controls how much of the document is constrained;
the two passes choose their sections independently. The released splits use $M \in \{3,5,7\}$ and $N \in \{2,3,4\}$ (structured)
or $\{3,5\}$ (flat); see Table~\ref{tab:pipeline-splits}.

\paragraph{Structure constraints.}
For each structured task, attach an exact-match lexical constraint on its title, and to every leaf an existence constraint requiring at least one paragraph of prose.
This supports scoring instructions like ``reproduce this exact outline''. The paragraph-existence check prevents a
model from satisfying the outline with empty sections. Flat tasks have no outline to reproduce, so the block is skipped.

\paragraph{Reward shaping.}
Each sampled constraint is compiled into an evaluation record: the checker configuration,
an output parser, and a weight group. The parser holds the path of the section the constraint
belongs to, optionally extended by an artifact tag and index, when the constraint is applied on a table/code-block or a list. Flat tasks do not receive a parser, so their checkers see the whole response.
Structure constraints are tagged \texttt{structure} and everything else \texttt{content},
and the task score is
$(1-w)\cdot\mathrm{mean}(\text{structure}) + w\cdot\mathrm{mean}(\text{content})$
with $w = \texttt{content\_weight} = 0.9$.
Flat tasks have no structure group and reduce to $\mathrm{mean}(\text{content})$.

\paragraph{Task verbalization.}
A verbalizer turns the outline and the sampled constraints into the user prompt. We
implemented several, so that the effect of prompt formatting can be measured separately from
the effect of the task itself. Structured splits are released with \texttt{default}: the
topic, the outline verbatim, a sentence requiring the headers to match it exactly, and one
bulleted block of constraints per section (Fig.~\ref{fig:example-task} in the main body). Flat splits are
released with \texttt{ifeval\_flat}, which drops the outline and the section attachment and
folds every constraint into running prose applying to the whole output, giving an
IFEval-style prompt. Two more verbalizers are implemented but unused in the release:
\texttt{inline\_constraints}, which interleaves each section's constraints into the outline
itself, and \texttt{flat}, a plain bulleted list. We also attempted an LLM-based verbalizer,
\texttt{verbalize\_topic}, which rewrites the templated opening into a more natural-looking
user request. It is not released: the rewriting model occasionally treated the task's
constraints as instructions addressed to itself and complied with them, returning a request
written in all caps, or in German with emoji, instead of a plain request \emph{for} a
document with those properties. We did not pursue it further, and suspect these pathologies
can largely be mitigated with better prompting.\footnote{Independently of which verbalizer
is used, \texttt{naturalize\_constraints} optionally asks an LLM to rephrase each
constraint's templated text so that it reads more naturally, preserving its meaning. It is
disabled in every released split.}

\paragraph{Splitting.}
The train/test split is taken on the stable \texttt{example\_id} key rather than by
position: a record is assigned to test if the MD5 digest of its key, read as a fraction in
$[0,1)$, falls below $\texttt{test\_size}/N$, where $\texttt{test\_size}=100$ and $N$ is the
size of the initial seed population.
A given topic therefore lands on the same side of the split in
every one of the ten configurations, no matter which items were dropped upstream by
generation failures.
This allows the splits to be pooled: a topic can appear in the
test set of several splits, but never in the test set of one and the training set of another.
We verify this property directly, by intersecting the union of all train keys with the union
of all test keys across the released ensemble.

\paragraph{Yield.}
All splits sharing a document setting are generated from one seed pool ($100$ test and
$3{,}900$ train topics), so the example counts in Table~\ref{tab:pipeline-splits} are also
retention rates: items that fail a stage are dropped entirely, and the failure rate grows
with outline size. On test, $10\%$ of topics are lost under \emph{flat}, $15\%$ under
\emph{short} and $34\%$ under \emph{long} ($2.3\%$, $5.6\%$ and $23.6\%$ on train).

\begin{table}[htbp]
  \centering
  \caption{The ten released splits and the pipeline parameters that distinguish them.
  \emph{sections} is $M$, the number of leaf sections one sampling pass may constrain;
  the two \emph{constraints / section} columns give $N$ per family, so a section receives
  between $1$ and $N$ constraints of that family. Every other pipeline parameter is
  identical across the ten splits, and the corresponding configuration keys are listed in
  Table~\ref{tab:pipeline-fixed}. Example counts are after failed items are dropped
  (see \emph{Yield}).}
  \label{tab:pipeline-splits}
  \small
  \setlength{\tabcolsep}{4pt}
  \begin{tabular}{llccccccc}
  \toprule
   &  & \multicolumn{2}{c}{\textbf{Document}} & \textbf{Scope} & \multicolumn{2}{c}{\textbf{Constraints / section}} & \multicolumn{2}{c}{\textbf{Examples}} \\
  \cmidrule(lr){3-4}\cmidrule(lr){5-5}\cmidrule(lr){6-7}\cmidrule(lr){8-9}
  \textbf{Split} & \textbf{Verbalization} & \textbf{branching} & \textbf{depth} & \textbf{sections} & \textbf{verifiable} & \textbf{judged} & \textbf{test} & \textbf{train} \\
  \midrule
  short / few      & structured & 3 & 2 & 3 & 1--2 & 1--2 & 85 & 3{,}681 \\
  short / some     & structured & 3 & 2 & 5 & 1--3 & 1--2 & 85 & 3{,}681 \\
  short / many     & structured & 3 & 2 & 7 & 1--3 & 1--3 & 85 & 3{,}681 \\
  short / numerous & structured & 3 & 2 & 7 & 1--4 & 1--4 & 85 & 3{,}681 \\
  \midrule
  long / few      & structured & 5 & 2 & 5 & 1--2 & 1--2 & 66 & 2{,}978 \\
  long / some     & structured & 5 & 2 & 7 & 1--3 & 1--2 & 66 & 2{,}978 \\
  long / many     & structured & 5 & 2 & 7 & 1--3 & 1--3 & 66 & 2{,}978 \\
  long / numerous & structured & 5 & 2 & 7 & 1--4 & 1--4 & 66 & 2{,}978 \\
  \midrule
  flat / some      & flat & --- & --- & 1 & 1--3 & 1--3 & 90 & 3{,}811 \\
  flat / many      & flat & --- & --- & 1 & 1--5 & 1--5 & 90 & 3{,}811 \\
  \bottomrule
  \end{tabular}
\end{table}

\begin{table}[htbp]
  \centering
  \caption{Pipeline hyper-parameters shared by all ten released splits; the four
  parameters that vary across splits are listed at the bottom, with their values in
  Table~\ref{tab:pipeline-splits}.}
  \label{tab:pipeline-fixed}
  \footnotesize
  \setlength{\tabcolsep}{5pt}
  \begin{tabular}{lll}
  \toprule
  \textbf{Stage} & \textbf{Parameter} & \textbf{Value} \\
  \midrule
  Seeding & seed corpus & Wikipedia Vital Articles, level 5 \\
          & \texttt{max\_examples} & 4000 \\
          & \texttt{random\_seed} & 1337 \\
  \midrule
  All LLM stages & model & Qwen2.5-32B-Instruct \\
          & temperature & 0.6 \\
          & \texttt{max\_tokens} & 2048 \\
  \midrule
  Outline & \texttt{document\_outline\_batch\_size} & 20 \\
          & \texttt{doc\_template\_verbalizer\_name} & \texttt{headers} \\
  \midrule
  Judged & \texttt{max\_content\_constraints} & 2 per section \\
         & \texttt{max\_artifact\_constraints} & 2 per section, per artifact type \\
         & \texttt{content\_eval\_type} / \texttt{artifact\_eval\_type} & \texttt{light\_critic} \\
         & judge $n$ / temp.\ / \texttt{max\_tokens} & 1 / 0.6 / 2048 \\
  \midrule
  Verifiable & keywords per section & 3 \\
         & \texttt{keyword\_generation\_max\_retries} & 3 \\
         & \texttt{use\_ifbench\_constraints} & \texttt{false} \\
  \midrule
  Sampling & \texttt{sample\_per\_header} & \texttt{true} \\
         & \texttt{exact\_sampling} & \texttt{false} \\
  \midrule
  Task & \texttt{use\_strict\_header\_names} & \texttt{true} (struct.), \texttt{false} (flat) \\
       & \texttt{content\_weight} & 0.9 (inert for flat) \\
       & \texttt{verbalize\_topic} / \texttt{naturalize\_constraints} & \texttt{false} \\
  \midrule
  Splitting & \texttt{split\_by\_field} & \texttt{example\_id} (hash split) \\
       & \texttt{test\_size} & 100 \\
  \midrule
  Varies by split & \texttt{doc\_template\_max\_children} / \texttt{\_max\_depth} & Table~\ref{tab:pipeline-splits} \\
       & \texttt{max\_sampled\_headers} & Table~\ref{tab:pipeline-splits} \\
       & \texttt{[non\_]verified\_max\_sampled\_constraints} & Table~\ref{tab:pipeline-splits} \\
  \midrule
  Infrastructure & \texttt{max\_concurrency} / \texttt{chunk\_size} & 20 / 500 \\
       & \texttt{min\_success\_threshold} & 0.2 \\
  \bottomrule
  \end{tabular}
\end{table}

%% file: appendix/judge_scoring.tex
\section{Judge Scoring}
\label{sec:appendix-judge}

This appendix describes how a judged constraint is turned into the boolean that enters a
task's score. Verifiable constraints bypass everything below: they are checked by
deterministic code and return a boolean directly.

\paragraph{Constraint types and checking mechanisms.}
Constraints are checked in one of two ways. \textbf{Verifiable} constraints use
deterministic checkers. \textbf{Judged} constraints are scored by an LM judge against criteria generated alongside the constraint.
Table~\ref{tab:constraint-types} gives the constraint types and the mechanism that checks each.

\begin{table}[htbp]
  \centering
  \caption{Constraint types used in the drafting pipeline and their scoring mechanisms, with examples.}
  \label{tab:constraint-types}
  \small
  \begin{tabular}{lll}
  \toprule
  \textbf{Type} & \textbf{Mechanism} & \textbf{Example} \\
  \midrule
  \texttt{doc\_structure}  & Verifiable    & Section headers must match the outline. \\[4pt]
  \texttt{language}        & Verifiable    & Write the text in Spanish. \\[4pt]
  \texttt{lexical\_count}  & Verifiable    & Use the word ``cards'' 3--6 times. \\[4pt]
  \texttt{emoji\_count}    & Verifiable    & Include at least 1 emoji. \\[4pt]
  \texttt{case\_type}      & Verifiable    & Write the text in uppercase. \\[4pt]
  \texttt{content}         & LM judge                & \parbox[t]{0.4\linewidth}{Discuss the historical significance of \textit{Dogs Playing Poker} in American pop culture.} \\[16pt]
  \texttt{tables}          & LM judge                & \parbox[t]{0.4\linewidth}{Compare lymphatic vs.\ blood vessel structure and function in a table.} \\[16pt]
  \texttt{list}            & LM judge                & \parbox[t]{0.4\linewidth}{List artists who created variations on the \textit{Dogs Playing Poker} theme.} \\[16pt]
  \texttt{code\_snippets}  & LM judge                & \parbox[t]{0.4\linewidth}{Include Python code computing reaction rates from a rate equation.} \\[16pt]
  \bottomrule
  \end{tabular}
  \end{table}

\paragraph{The light critic.}
Judged constraints are checked by a \emph{light critic}.
The judge is shown a reconstructed request, the criterion under test, and the model's
text for that scope alone rather than the whole document, and returns an integer on a
$1$--$5$ scale inside \texttt{<decision>} tags. The integer is
converted into a score using a threshold $int(score \ge 4)$. The judge is run with $n=1$ at temperature
$0.6$, using Gemma-4-31B as the LM judge. It is not fine-tuned
for this task, and the threshold applied to the rubric score is not calibrated against labelled judgments. Appendix~\ref{sec:appendix-judge-meta} measures the judge's agreement
with human annotators and ablates the scoping, the judge model and the prompt.

\paragraph{Judge prompt.}
The judge is queried with the template below, once per judged constraint.
\texttt{\{request\}} is \emph{not} the prompt the evaluated model saw: it is a synthetic
single-section request written for the judge (``I'm writing a document about
\emph{topic}. Write content for the section \emph{section}.''), which tells the judge what
the text was supposed to be without restating any of the task's constraints.
\texttt{\{response\}} is the model's text for that scope alone, and
\texttt{\{criterion\}} the single constraint under test, so the judge is only asked
whether one scope satisfies one constraint. The scoping is deliberate: the judge is handed
exactly the text a constraint applies to, instead of requiring the model to search for the relevant passage inside a full document that can potentially be very long.

\begin{footnotesize}
\begin{verbatim}
Denote the following request and response, enclosed by xml tags
<request>REQUEST</request> and <response>RESPONSE</response> respectively.

<request>
{request}
</request>

<response>
{response}
</response>

Based ONLY on the given details, answer the question: Does the response 
adhere to the following criterion:
{criterion}

Your answer should be a score on a scale of 1 to 5, and should be
formatted in XML tags as follows: <decision>ANSWER</decision>.
Provide only the final answer, without additional explanations.

If the criterion is not applicable to the request and response, answer
5.
\end{verbatim}
\end{footnotesize}

%% file: appendix/judge_meta_eval.tex
\section{Judge Meta-Evaluation}
\label{sec:appendix-judge-meta}

Appendix~\ref{sec:appendix-judge} describes how the LM judge is queried; this appendix evaluates its correctness, against human labels.

\paragraph{Annotated set.}
We sampled $305$ \emph{(constraint, response)} pairs from judged constraints on
\benchmark{}, stratified over the judged constraint types of
Table~\ref{tab:constraint-types} and over the baseline judge's own verdict.
Human annotators then labelled each pair from the same scope and criterion the judge sees.
Of the $305$ pairs, $227$ are \emph{valid} and $78$ \emph{invalid}; we report macro-$F_1$
over the two classes and the raw confusion counts.

\paragraph{Configurations.}
The headline comparison is between two settings, run over several judge models.
\emph{Unscoped} gives the judge the original user request - the full task prompt with all
of its constraints - plus the model's complete response, and asks about one criterion;
this is the conventional LM-as-a-judge setup, and what a benchmark gets by default when
constraints are not tracked back to the text they govern. \emph{Scoped} gives it the
synthetic single-section request of Appendix~\ref{sec:appendix-judge} plus only the
artifact the criterion applies to, and is what \benchmark{} ships with.

\begin{table}[htbp]
\centering
\caption{Judge meta-evaluation against human labels ($n=305$; $227$ valid, $78$ invalid).
The original baseline judge is marked $\dagger$. Every scoped judge beats every unscoped one, regardless of model.}
\label{tab:judge-meta-eval}
\small
\setlength{\tabcolsep}{8pt}
\begin{subtable}{\linewidth}
\centering
\caption{\emph{Unscoped}: original request $+$ full response}
\label{tab:judge-meta-eval-unscoped}
\begin{tabular}{@{}p{2.6cm}rrrrc@{}}
\toprule
\textbf{Judge model} & \textbf{TP} & \textbf{FP} & \textbf{TN} & \textbf{FN} & \textbf{Macro-}$\mathbf{F_1}$ \\
\midrule
Qwen2.5-32B & 205 & 74 &  4 & 22 & $0.444$ \\
Gemma-4-31B & 224 & 68 & 10 &  3 & $0.541$ \\
\bottomrule
\end{tabular}
\end{subtable}

\vspace{8pt}
\begin{subtable}{\linewidth}
\centering
\caption{\emph{Scoped}: synthetic section request $+$ governed artifact}
\label{tab:judge-meta-eval-scoped}
\begin{tabular}{@{}p{2.6cm}rrrrc@{}}
\toprule
\textbf{Judge model} & \textbf{TP} & \textbf{FP} & \textbf{TN} & \textbf{FN} & \textbf{Macro-}$\mathbf{F_1}$ \\
\midrule
Qwen2.5-32B$^\dagger$ & 131 & 29 & 49 & 96 & $0.558$ \\
Qwen3-8B              & 188 & 50 & 28 & 39 & $0.597$ \\
Qwen3-32B             & 155 & 30 & 48 & 72 & $0.619$ \\
Gemma-4-31B           & 221 & 45 & 33 &  6 & $\mathbf{0.730}$ \\
\bottomrule
\end{tabular}
\end{subtable}
\end{table}

\paragraph{Scoping is what makes the judge work.}
Holding the model fixed, scoping is worth roughly $11$--$19$ macro-$F_1$ points depending
on the judge --- a larger effect than any other axis we varied, and large enough that the
weakest scoped judge beats the strongest unscoped one. When asked
about a whole document, both models accept almost everything: Qwen2.5-32B answers
\emph{valid} for $279$ of $305$ pairs, and Gemma-4-31B for $292$ of $305$ pairs.
Both become discriminative only once the criterion and the text it governs are
handed over together. A judge given a long document and a single criterion mostly fails at
locating the relevant passage, not at judging it --- which is the empirical content of the
scoping argument in \S\ref{sec:draftingbench-task}.

\paragraph{The judge model matters, but less than the scope.}
Within the scoped setting, the alternatives to the Qwen2.5-32B baseline are worth between
$4$ and $17$ points, and we adopt the strongest, Gemma-4-31B~\citep{gemmateam2026gemma4}, at comparable
serving cost.
The lever is real but weaker and less reliable than scoping.

\paragraph{Further ablations.}
We ran two further ablations within the scoped setting: varying the scoping level, passing
the enclosing section rather than the judged artifact alone, and including the grading rubric
in the judge prompt. Both move macro-$F_1$ by at most a point over the configuration we
adopt. The remaining gap to human agreement is therefore large and unlikely to close by
prompting alone; a trained judge is the more promising route, and
\citet{he2026advancedif} give a recipe that consumes labels of exactly the kind collected
here, though it would need a set considerably larger than ours. Calibrating the hand-set
$1$--$5$ threshold of Appendix~\ref{sec:appendix-judge} against these labels is a cheaper
first step.

%% file: appendix/splits.tex
\section{Statistics for Released Splits}
\label{sec:splits-stats}

\begin{figure}[htbp]
\centering
\begin{subfigure}[t]{\textwidth}
    \centering
    \includegraphics[width=\textwidth]{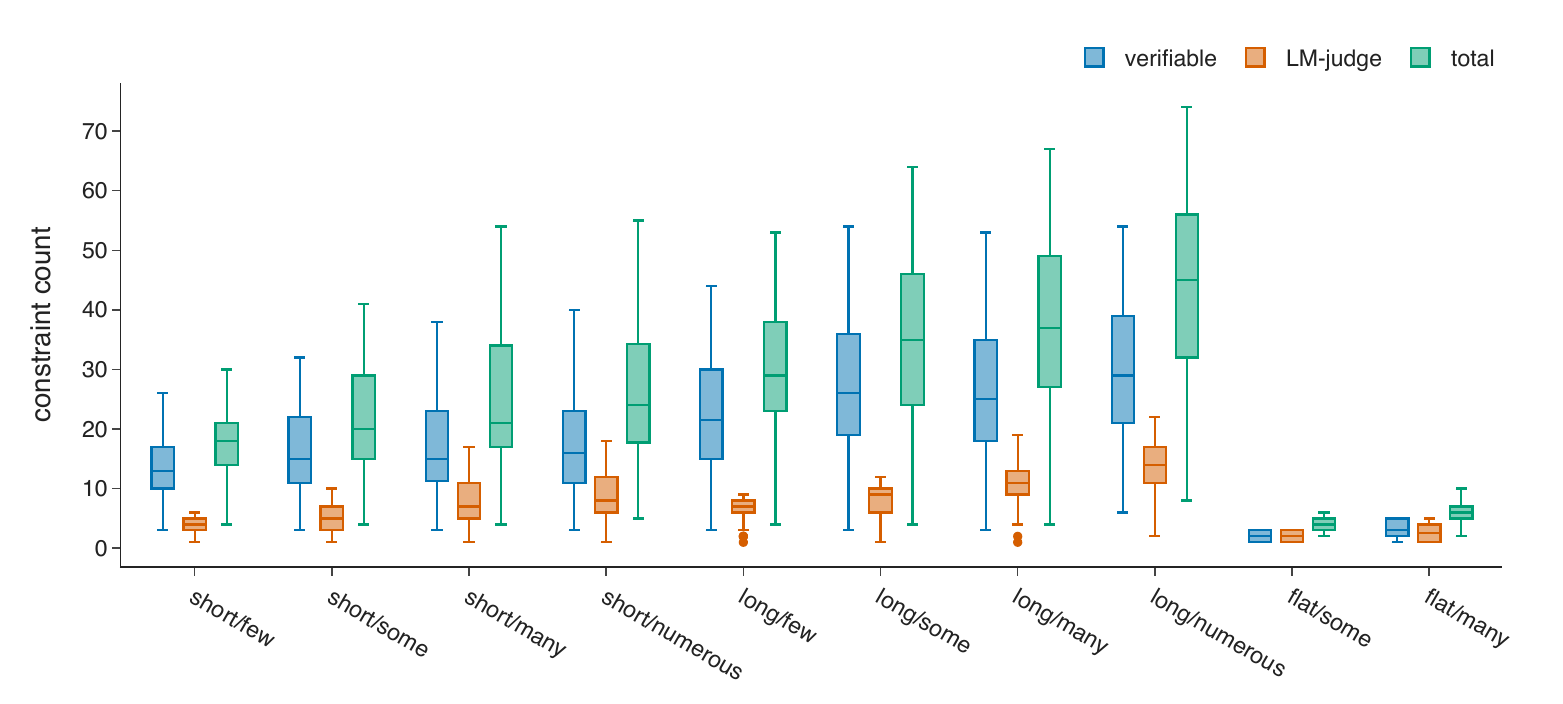}
    \caption{Constraints per example, by split: verifiable, judged (LM-judge) and total.}
    \label{fig:constraint-counts}
\end{subfigure}

\vspace{0.8em}

\begin{subfigure}[t]{\textwidth}
    \centering
    \includegraphics[width=\textwidth]{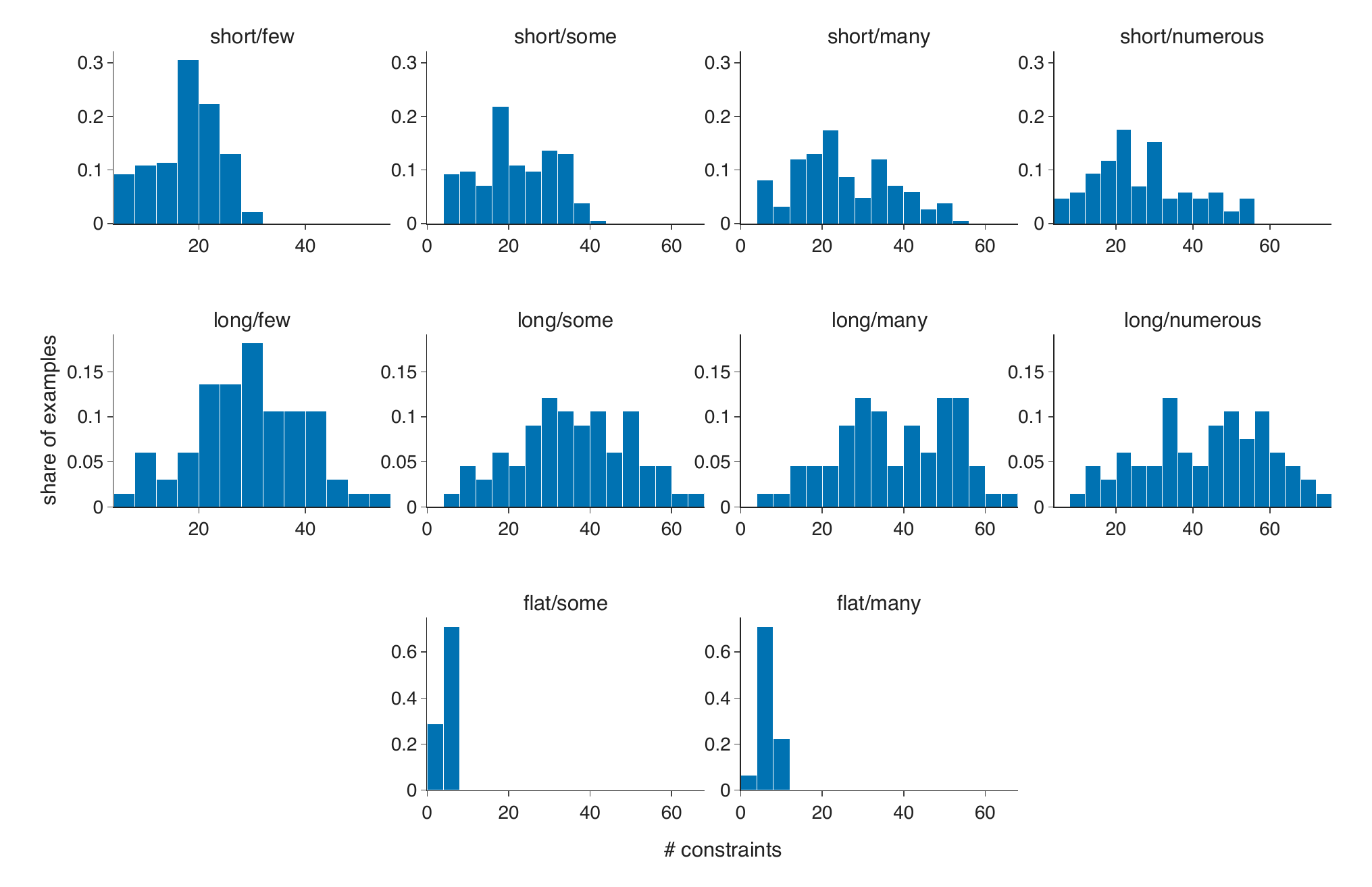}
    \caption{Constraint-count distribution per split, one histogram per split.}
    \label{fig:difficulty-mix}
\end{subfigure}
\caption{Descriptive statistics for the released splits.}
\label{fig:splits-identity-card}
\end{figure}

Figure~\ref{fig:splits-identity-card} provides some descriptive statistics for the released splits, showing the number of constraints per example, by split, and the composition of difficulty buckets per split.

%% file: appendix/models.tex
\section{Evaluated Models}
\label{sec:appendix-models}

\S\ref{sec:analysis-setup} names the nine models reported in
Table~\ref{tab:leaderboard}. This appendix gives the complete set of 17 evaluated models,
with citations.

\paragraph{Closed-weight.}
GPT-5.5~\citep{openai2026gpt55}, GPT-5.6 Sol and GPT-5.6 Luna~\citep{openai2026gpt56},
GPT-5 nano~\citep{openai2026gpt5card}, GPT-5.4 nano~\citep{openai2026gpt54nano}, Claude
Opus 5~\citep{anthropic2026opus5}, Claude Fable 5~\citep{anthropic2026fable5}, Claude Opus
4.8~\citep{anthropic2026opus48}, Claude Sonnet 4.6~\citep{anthropic2026sonnet46}, and
Claude Haiku 4.5~\citep{anthropic2025haiku45}.

\paragraph{Open-weight.}
DeepSeek-V4-Pro~\citep{deepseek2026v4}, GLM-5.2 and GLM-5.3~\citep{glm2026glm5}, Kimi K3~\citep{kimi2026k3},
gpt-oss-20b~\citep{openai2025gptoss}, Qwen3-32B~\citep{yang2025qwen3}, and
Qwen3.8-2.4T-A95B~\citep{qwen2026qwen38}.

%% file: appendix/full_results.tex
\section{Full Results}
\label{sec:appendix-full-results}

Table~\ref{tab:leaderboard} in \S\ref{sec:analysis-overall} reports selected
models. Table~\ref{tab:leaderboard-full} gives per-split scores for all 17 evaluated
models (\S\ref{sec:appendix-models}).

\begin{table}[htbp]
\centering
\caption{Mean score per split for all 17 evaluated models, shown as two panels: the ten closed-weight models in~(\subref{tab:leaderboard-full-closed})
and the seven open-weight models in~(\subref{tab:leaderboard-full-open}). \textbf{Bold marks the best value across all 17 models, not
the best within a panel.} Averages carry a $\pm$95\% CI across examples; `avg (all)' weighs the structured and flat averages equally.}
\label{tab:leaderboard-full}

\begin{subtable}[t]{\textwidth}
  \centering
  \caption{closed-weight models}
  \label{tab:leaderboard-full-closed}
  \resizebox{\textwidth}{!}{\input{figures/benchmark_analysis/leaderboard_full_closed.tex}}
\end{subtable}

\vspace{1.5em}

\begin{subtable}[t]{\textwidth}
  \centering
  \caption{open-weight models}
  \label{tab:leaderboard-full-open}
  \resizebox{\textwidth}{!}{\input{figures/benchmark_analysis/leaderboard_full_open.tex}}
\end{subtable}
\end{table}

%% file: figures/benchmark_analysis/leaderboard_full_closed.tex
\begin{tabular}{lcccccccccc}
\toprule
 & \multicolumn{5}{c}{GPT} & \multicolumn{5}{c}{Claude} \\
\cmidrule(lr){2-6} \cmidrule(lr){7-11}
split & 5.5 & 5.6 Sol & 5.6 Luna & 5 nano & 5.4 nano & Opus 5 & Fable 5 & Opus 4.8 & Sonnet 4.6 & Haiku 4.5 \\
\midrule
short/few & \textbf{0.97} & 0.95 & 0.88 & 0.71 & 0.68 & 0.92 & 0.92 & 0.94 & 0.76 & 0.71 \\
short/some & \textbf{0.96} & 0.94 & 0.84 & 0.62 & 0.66 & 0.88 & 0.91 & 0.92 & 0.78 & 0.73 \\
short/many & \textbf{0.98} & 0.94 & 0.78 & 0.58 & 0.60 & 0.84 & 0.82 & 0.89 & 0.68 & 0.65 \\
short/numerous & \textbf{0.93} & 0.92 & 0.71 & 0.47 & 0.57 & 0.81 & 0.81 & 0.82 & 0.62 & 0.57 \\
long/few & \textbf{0.98} & 0.96 & 0.89 & 0.67 & 0.59 & 0.91 & 0.92 & 0.96 & 0.79 & 0.79 \\
long/some & \textbf{0.95} & 0.95 & 0.83 & 0.61 & 0.59 & 0.93 & 0.91 & 0.91 & 0.76 & 0.75 \\
long/many & \textbf{0.97} & 0.93 & 0.77 & 0.55 & 0.53 & 0.89 & 0.90 & 0.92 & 0.70 & 0.67 \\
long/numerous & 0.91 & 0.89 & 0.66 & 0.47 & 0.53 & 0.84 & 0.83 & 0.87 & 0.64 & 0.60 \\
flat/some & 0.71 & 0.66 & 0.63 & 0.73 & 0.57 & 0.65 & 0.64 & 0.71 & 0.59 & 0.58 \\
flat/many & 0.72 & 0.65 & 0.63 & 0.68 & 0.59 & 0.65 & 0.66 & 0.74 & 0.59 & 0.56 \\
\midrule
avg (structured) & \textbf{0.96}{\scriptsize$\pm$0.01} & 0.94{\scriptsize$\pm$0.01} & 0.80{\scriptsize$\pm$0.02} & 0.59{\scriptsize$\pm$0.03} & 0.60{\scriptsize$\pm$0.02} & 0.87{\scriptsize$\pm$0.01} & 0.88{\scriptsize$\pm$0.01} & 0.90{\scriptsize$\pm$0.01} & 0.72{\scriptsize$\pm$0.01} & 0.68{\scriptsize$\pm$0.02} \\
avg (flat) & 0.71{\scriptsize$\pm$0.03} & 0.66{\scriptsize$\pm$0.03} & 0.63{\scriptsize$\pm$0.03} & 0.71{\scriptsize$\pm$0.04} & 0.58{\scriptsize$\pm$0.03} & 0.65{\scriptsize$\pm$0.03} & 0.65{\scriptsize$\pm$0.03} & 0.72{\scriptsize$\pm$0.03} & 0.59{\scriptsize$\pm$0.03} & 0.57{\scriptsize$\pm$0.03} \\
avg (all) & 0.84{\scriptsize$\pm$0.02} & 0.80{\scriptsize$\pm$0.01} & 0.72{\scriptsize$\pm$0.02} & 0.65{\scriptsize$\pm$0.02} & 0.59{\scriptsize$\pm$0.02} & 0.76{\scriptsize$\pm$0.02} & 0.76{\scriptsize$\pm$0.02} & 0.81{\scriptsize$\pm$0.02} & 0.65{\scriptsize$\pm$0.02} & 0.63{\scriptsize$\pm$0.02} \\
\bottomrule
\end{tabular}

%% file: figures/benchmark_analysis/leaderboard_full_open.tex
\begin{tabular}{lccccccc}
\toprule
 & \multicolumn{1}{c}{DeepSeek} & \multicolumn{2}{c}{GLM} & \multicolumn{1}{c}{Kimi} & \multicolumn{1}{c}{GPT} & \multicolumn{2}{c}{Qwen} \\
\cmidrule(lr){2-2} \cmidrule(lr){3-4} \cmidrule(lr){5-5} \cmidrule(lr){6-6} \cmidrule(lr){7-8}
split & V4 Pro & 5.2 & 5.3 & K3 & OSS-20B & Qwen3-32B & Qwen3.8-A95B \\
\midrule
short/few & 0.95 & 0.92 & 0.87 & 0.97 & 0.84 & 0.75 & 0.89 \\
short/some & 0.95 & 0.89 & 0.88 & 0.96 & 0.81 & 0.72 & 0.88 \\
short/many & 0.93 & 0.86 & 0.83 & 0.97 & 0.75 & 0.65 & 0.82 \\
short/numerous & 0.92 & 0.85 & 0.80 & 0.92 & 0.71 & 0.58 & 0.81 \\
long/few & 0.95 & 0.89 & 0.93 & 0.96 & 0.82 & 0.78 & 0.90 \\
long/some & 0.94 & 0.86 & 0.89 & 0.95 & 0.78 & 0.74 & 0.89 \\
long/many & 0.95 & 0.83 & 0.87 & 0.95 & 0.71 & 0.69 & 0.87 \\
long/numerous & 0.90 & 0.71 & 0.84 & \textbf{0.92} & 0.68 & 0.61 & 0.80 \\
flat/some & 0.82 & 0.75 & 0.60 & \textbf{0.85} & 0.67 & 0.59 & 0.71 \\
flat/many & 0.80 & 0.81 & 0.60 & \textbf{0.83} & 0.67 & 0.62 & 0.76 \\
\midrule
avg (structured) & 0.94{\scriptsize$\pm$0.01} & 0.86{\scriptsize$\pm$0.02} & 0.86{\scriptsize$\pm$0.01} & 0.95{\scriptsize$\pm$0.01} & 0.76{\scriptsize$\pm$0.01} & 0.69{\scriptsize$\pm$0.01} & 0.86{\scriptsize$\pm$0.01} \\
avg (flat) & 0.81{\scriptsize$\pm$0.03} & 0.78{\scriptsize$\pm$0.04} & 0.60{\scriptsize$\pm$0.03} & \textbf{0.84}{\scriptsize$\pm$0.03} & 0.67{\scriptsize$\pm$0.03} & 0.60{\scriptsize$\pm$0.03} & 0.73{\scriptsize$\pm$0.03} \\
avg (all) & 0.87{\scriptsize$\pm$0.02} & 0.82{\scriptsize$\pm$0.02} & 0.73{\scriptsize$\pm$0.01} & \textbf{0.90}{\scriptsize$\pm$0.02} & 0.72{\scriptsize$\pm$0.02} & 0.65{\scriptsize$\pm$0.02} & 0.79{\scriptsize$\pm$0.02} \\
\bottomrule
\end{tabular}

%% file: appendix/flat_anomaly.tex
\section{Globally-Scoped Prompts Are the Hardest Regime}
\label{sec:appendix-flat-anomaly}

\begin{figure}[htbp]
\centering
\includegraphics[width=\textwidth]{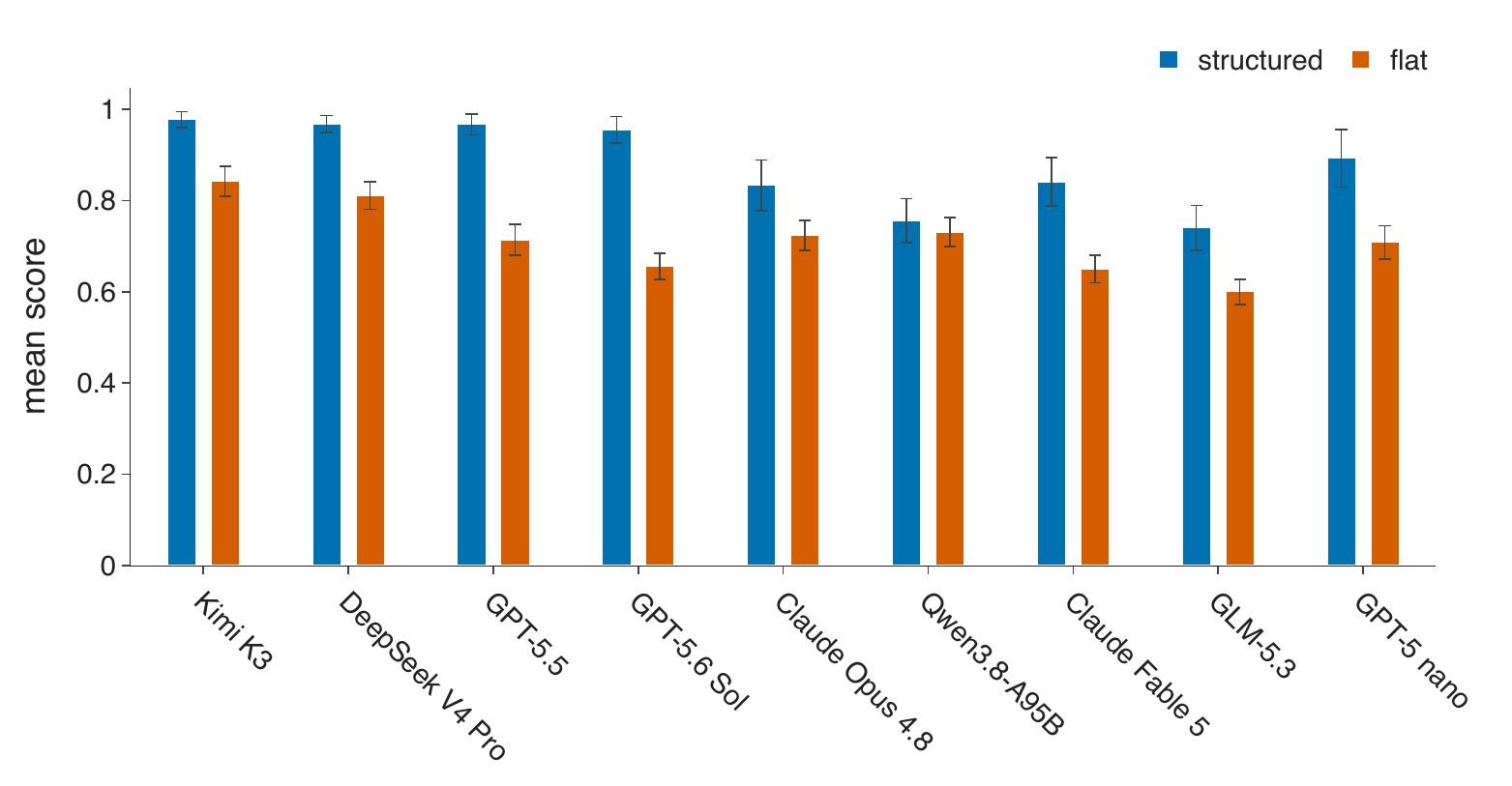}
\caption{Fewer constraints, worse scores: the globally-scoped (flat) regime is harder
than the scoped one and ranks models differently. Mean score ($\pm$95\% CI) at a matched
constraint budget ($\leq$10 total constraints), structured vs.\ flat}
\label{fig:flat-anomaly}
\end{figure}

Flat tasks carry an order of magnitude fewer constraints
(\S\ref{sec:draftingbench-splits}), yet they are the frontier models'
\emph{worst} cells: the leading closed models drop to 0.65--0.74, below all of their
structured cells. This drop cannot be explained merely by constraint count, as restricting both scopes to examples with $\leq$10 total constraints
(Fig.~\ref{fig:flat-anomaly}), all but one model (Qwen3.8-A95B) scores significantly
lower on flat than on structured tasks.

%% file: appendix/full_heatmaps.tex
\section{Full Model Set}
\label{sec:appendix-full-heatmaps}

Figure~\ref{fig:type-fingerprint} and Table~\ref{tab:section-load} show five models spanning
the capability range. This appendix gives both over a wider set: the nine evaluated models plus eight smaller ones from earlier rounds.

Frontier models are largely indistinguishable here, and the variance comes from the smaller
ones. Showing five models is a presentation choice, not a filter on the evidence.

\begin{figure}[htbp]
\centering
\includegraphics[width=\textwidth]{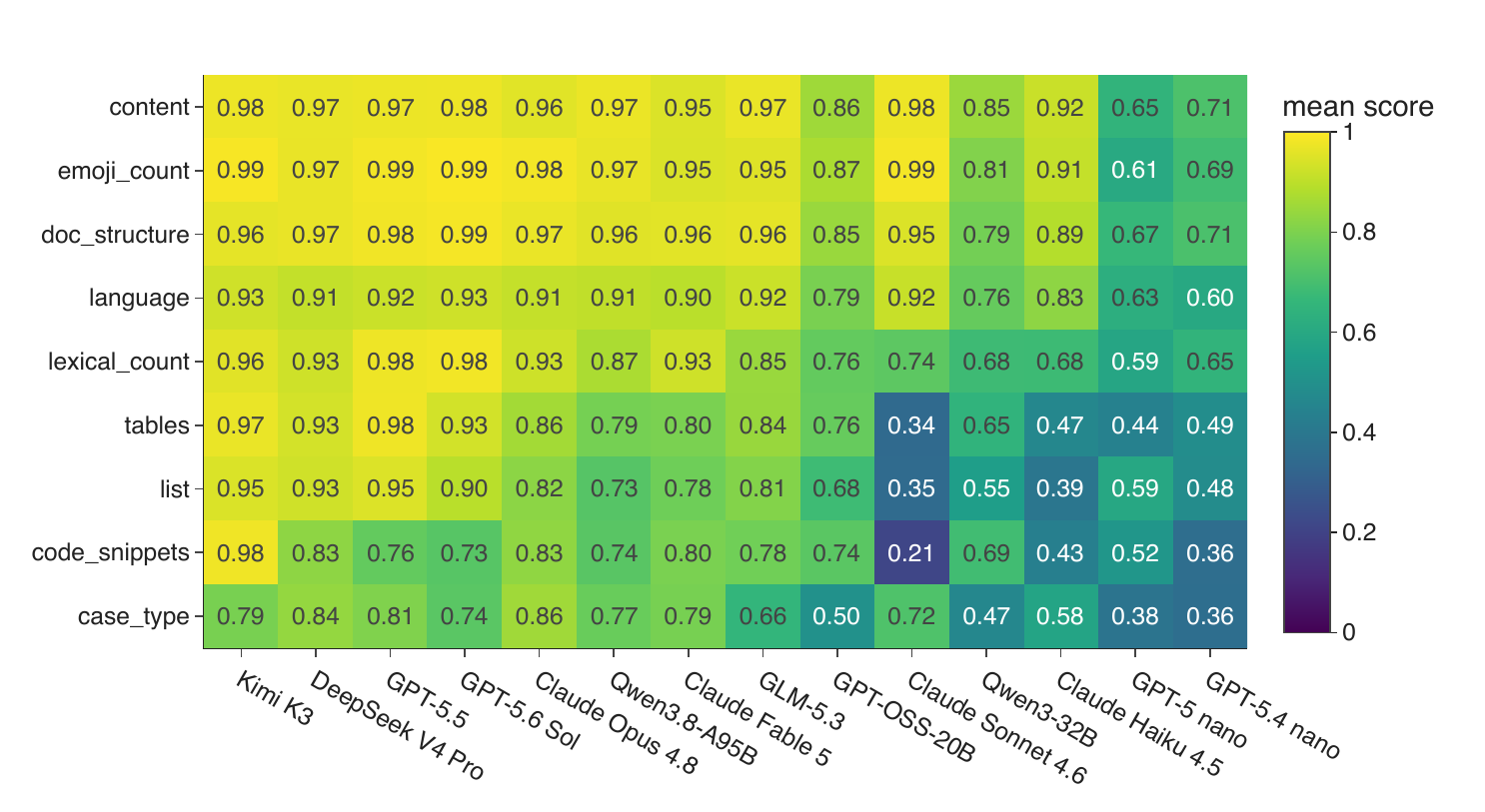}
\caption{Constraint-type fingerprint, structured splits pooled.}
\label{fig:type-fingerprint-all}
\end{figure}

\begin{table}[htbp]
\centering
\caption{Section-load slopes. Columns as in Table~\ref{tab:section-load}.}
\label{tab:section-load-all}
\input{figures/benchmark_analysis/section_load_slopes_all_models.tex}
\end{table}

%% file: figures/benchmark_analysis/section_load_slopes_all_models.tex
\begin{tabular}{lrrr}
\toprule
 & \multicolumn{2}{c}{\% score change by document size} & \\
\cmidrule(lr){2-3}
Model & short & long & mean \\
\midrule
Kimi K3 & $-0.9$\% & $+0.2$\% & $-0.4$\% \\
Claude Opus 4.8 & $-0.4$\% & $-0.5$\% & $-0.4$\% \\
DeepSeek V4 Pro & $-0.8$\% & $-0.5$\% & $-0.7$\% \\
GPT-5.6 Sol & $-0.4$\% & $-0.9$\% & $-0.7$\% \\
Qwen3.8-A95B & $-1.1$\% & $-1.2$\% & $-1.1$\% \\
GLM-5.3 & $-1.3$\% & $-1.1$\% & $-1.2$\% \\
Claude Sonnet 4.6 & $-1.2$\% & $-1.4$\% & $-1.3$\% \\
GPT-5.5 & $-1.3$\% & $-1.7$\% & $-1.5$\% \\
Claude Fable 5 & $-3.5$\% & $-0.0$\% & $-1.7$\% \\
GPT-5.4 nano & $-4.1$\% & $-1.4$\% & $-2.7$\% \\
Claude Haiku 4.5 & $-4.0$\% & $-3.0$\% & $-3.5$\% \\
GPT-OSS-20B & $-3.7$\% & $-3.9$\% & $-3.8$\% \\
Qwen3-32B & $-4.3$\% & $-5.1$\% & $-4.7$\% \\
GPT-5 nano & $-7.0$\% & $-4.1$\% & $-5.5$\% \\
\bottomrule
\end{tabular}

%% file: appendix/section_load_by_setting.tex
\section{Section-Load Fitting}
\label{sec:appendix-section-load}

Table~\ref{tab:section-load} reports, per model, how much a section's score changes per
density step (few$\to$numerous), which adds constraints
\emph{elsewhere} in the document while the section's own constraint count is unchanged. This
appendix gives the procedure.

Each slope is an ordinary least-squares fit of section score on density tier (few$=0$,
some$=1$, many$=2$, numerous$=3$), over structured splits only --- flat tasks have a single
global scope and no notion of ``elsewhere in the document.'' The unit of observation is one
section instance, and a section's score is the mean over the constraints anchored to it.

Slopes are fitted separately within the \emph{short} and \emph{long} document settings, as the scores in each setting change a bit differently.
 \emph{Short} documents carry too few sections at per-section loads $2$ and $3$ in the denser tiers
(Table~\ref{tab:pipeline-splits}) and admit no fit there at all.

Section's constraint count ranges between $1$ and $6$ and is highly imbalanced,
so the slopes are stratified by section's constraint count and averaged per model within each
setting; the \emph{mean} column of Table~\ref{tab:section-load} is the average of the
\emph{short} and \emph{long} values.

%% file: appendix/reward_hacking_analysis.tex
\section{Reward Hacking Analysis}
\label{app:reward-hacking}

Figure \ref{fig:reward-hacking} shows an analysis of reward hacking behavior, demonstrating that training on IFTrain leads to short, degenerate outputs, which optimize constraint score but do not adhere to the prompt content.

\input{figures/training_experiments/reward_hacking.tex}

\subsection{Reward-hacked completions}
\label{app:reward-hacking-examples}

The transcripts below are reproduced from held-out validation passes of the IFTrain run. IFTrain's reward is a lexical instruction-following
checker: it returns $1.0$ when every constraint attached to the prompt is satisfied and
$1/256$ otherwise, and it never inspects whether the response answers the question.
Every completion shown scores the full $1.0$. Completions are verbatim.

{\small\begin{verbatim}
Prompt: Include quotes within quotes within quotes, at least 3 levels 
deep, alternating between double quotes and single quotes. How many
solutions are there to the equation $x_1 + x_2 + x_3 + x_4 = 20$,
where $x_i$ is a non-negative integer, for all $i$, if (a) there are
no restrictions? (b) $x_1 > 1$, $x_2 > 2$, $x_3 > 3$, $x_4 > 4$? (c)
$x_1 > 4$ and $x_3 < 5$?

Completion: "'"10"'""5""'6'"""2""'"3'"""1""'"4"""'"5""'"6"""'"7""'"8""
"'"9""'"10"""'"11""'"12"""'"13""'"14"""'"15""'"16"""'"17""'"18"""'"19"
"'"20"""'"21""'"22""'" [... the same pattern, ascending to "427]
\end{verbatim}}

{\small\begin{verbatim}
Prompt: [a prompt requesting a chapter of an alternate history of the
1911--1912 Xinhai Revolution, with no word repeated more than four times]

Completion:  China
\end{verbatim}}

%% file: figures/training_experiments/reward_hacking.tex
\providecommand{\benchmarkTrain}{LIFT-Train}
\begin{figure}[htbp]
\centering
\begin{subfigure}[t]{0.49\textwidth}
  \centering
  \includegraphics[width=\textwidth]{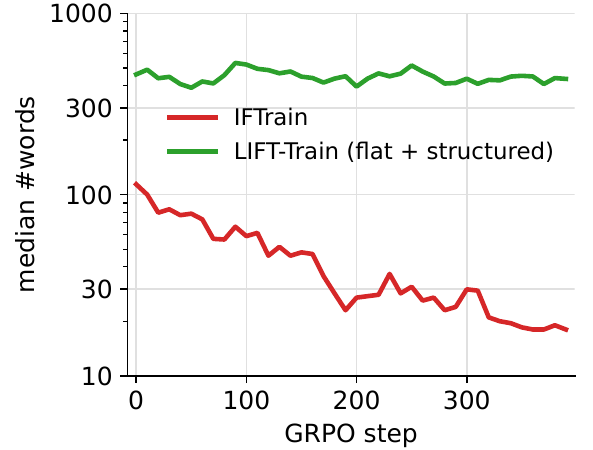}
  \caption{Median visible answer length, log scale. IFTrain collapses from $114$ words to $28.5$ by step $180$ and $18$ by step $390$, while \benchmarkTrain{} stays between $389$ and $533.5$.}
  \label{fig:reward-hacking-length}
\end{subfigure}
\hfill
\begin{subfigure}[t]{0.49\textwidth}
  \centering
  \includegraphics[width=\textwidth]{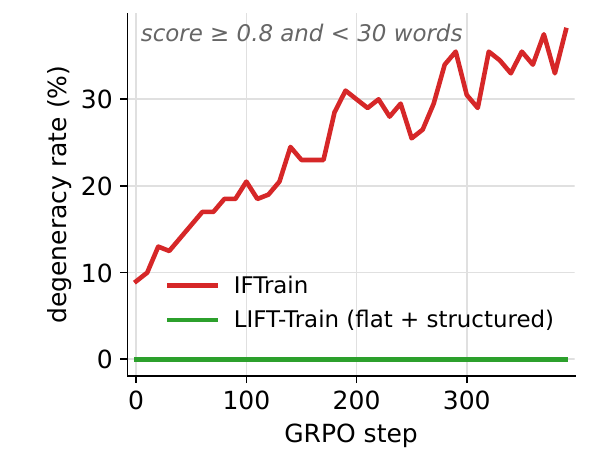}
  \caption{Share of completions scoring $\geq 0.8$ with fewer than $30$ words. IFTrain reaches $28.5\%$ at step $180$ and is still rising at $38\%$ by step $390$; \benchmarkTrain{} stays at $0\%$ throughout.}
  \label{fig:reward-hacking-degeneracy}
\end{subfigure}
\caption{Rollout health during GRPO training on Qwen2.5-7B: the lexical-only reward is the one that gets hacked. Held-out validation pass every $10$ steps ($200$ completions per point).}
\label{fig:reward-hacking}
\end{figure}